\documentclass[letterpaper]{article} %
\usepackage{aaai2026}  %
\usepackage{times}  %
\usepackage{helvet}  %
\usepackage{courier}  %
\usepackage[hyphens]{url}  %
\usepackage{graphicx} %
\usepackage{natbib}  %
\usepackage{caption} %
\usepackage{algorithm}
\usepackage[noend]{algpseudocode}
\usepackage{amsmath}

\usepackage{newfloat}
\usepackage{listings}
\DeclareCaptionStyle{ruled}{labelfont=normalfont,labelsep=colon,strut=off} %
\floatstyle{ruled}
\newfloat{listing}{tb}{lst}{}
\floatname{listing}{Listing}
\title{Evaluation of Baseline Methods for IDD-based SSD External Memory Search\thanks{Accepted to the International Symposium on Combinatorial Search (SoCS 2026). This preprint version includes an Appendix with additional details and results.}}
\author{
    Yuki Suzuki,
    Alex Fukunaga
}
\affiliations{
    The University of Tokyo\\
    suzuki-yuki0631@g.ecc.u-tokyo.ac.jp, fukunaga@idea.c.u-tokyo.ac.jp
}

\usepackage{bibentry}

\def\arxiv{}  %

\usepackage{comment}
\usepackage{booktabs}
\usepackage{subcaption}
\usepackage{xspace}
\usepackage{xcolor}
\usepackage{placeins}
\usepackage{amsthm}
\usepackage{colortbl} %
\usepackage{inconsolata} %

\newcommand{\astar}{$\mathit{A^*}$\xspace}
\newcommand{\siddastar}{SIDD-$\mathit{A^*}$\xspace}
\newcommand{\siddastarpwrite}{SIDD-$\mathit{A^*_{p}}$\xspace}
\newcommand{\siddastarmmap}{SIDD-$\mathit{A^*_{m}}$\xspace}

\newcommand{\open}{\textsc{Open}\xspace}
\newcommand{\closed}{\textsc{Closed}\xspace}

\newcommand{\pddl}[1]{\textsf{\small #1}\xspace}

\newcommand{\brd}{\texttt{brd}\xspace}
\newcommand{\tmpfs}{\texttt{tmpfs}\xspace}
\newcommand{\ssd}{\texttt{ssd}\xspace}

\newcolumntype{G}{!{\color{gray!40}\vrule}} %

\theoremstyle{definition}
\newtheorem{observation}{Observation}

\def\verifiedexpansions{}  %

\ifdefined\arxiv
\nocopyright

\else
\usepackage{xr-hyper}
\fi

\begin{document}

\maketitle

\begin{abstract}

Many difficult search problems  cannot be solved by algorithms such as \astar using only RAM. Search algorithms which use external memory such as SSDs and HDDs with much higher capacity than RAM have been proposed in previous work, but previous work has focused on delayed duplicate detection approaches,  as well as complex immediate duplicate detection (IDD) methods, and relatively simple methods for IDD have not been systematically studied. In addition, the effect of OS-level mechanisms for managing and speeding up accesses to external memory, such as page caches, has not been studied.
This paper addresses these gaps in the literature by evaluating and analyzing the performance of simple baseline approaches for IDD-based \astar.

\end{abstract}

\section{Introduction}

Heuristic search algorithms such as \astar are limited by the memory available to store nodes. One approach to overcoming memory limitations is to use external memory such as hard disk drives (HDDs) or solid state drives (SSDs). Since accessing external memory is significantly slower than accessing RAM, previous work on search algorithms investigated methods to overcome this performance bottleneck. 

Early work on external-memory search focused on Delayed Duplicate Detection (DDD) methods that reorganized the structure of the search algorithm in order to replace random accesses with sequential reads and writes, an approach which was especially effective in overcoming the massive (6 orders of magnitude) latency difference between RAM and HDDs.
More recently, the ubiquity of SSDs, which are 3 orders of magnitude faster than HDDs, made Immediate Duplicate Detection (IDD), which allows standard search strategies (node expansion orders) to be easily implemented, a practical alternative to DDD.

Although SSDs are much faster than HDDs, they are orders of magnitude slower than RAM, so algorithms that use SSD as external memory were designed to reduce accesses to the SSD.
Early work on external-memory search for model checking showed that 
multi-layer data structures which partly reside in RAM and partly on the SSD are effective in improving performance by reducing SSD accesses \cite{EDELKAMP2011136} . A*-IDD \cite{DBLP:conf/aaai/LinF18} is an implementation of \astar with IDD, which uses {\it compression}, a multi-layer technique for implementing the \closed hash table in order to reduce the cost of duplicate detection \cite{EDELKAMP2011136}. 
Compression seeks to minimize read accesses to the SSD by implementing \closed as a two-level hash table, where an internal hash table in RAM points to entries in the external hash table on the SSD. 
A*-IDD was shown to outperform DDD-based \astar on IPC domain-independent planning benchmarks.

However, IDD-based external-memory \astar is not well understood. 
Although it was shown that A*-IDD, a relatively complicated algorithm, could be competitive with DDD approaches, the factors underlying the performance of A*-IDD were not investigated in depth. For example, it was assumed that multi-layered techniques such as compression were necessary,   and Lin and Fukunaga (\citeyear{DBLP:conf/aaai/LinF18}) compared two variants of compression \cite{EDELKAMP2011136}, where RAM is used to store a compressed index of pointers to the \closed data structure on external memory, but did not evaluate simpler, baseline implementations of IDD. 
As another example, Lin and Fukunaga (\citeyear{DBLP:conf/aaai/LinF18}) allude to the significant effects of the OS page cache mechanism on overall algorithm performance, but to our knowledge, previous work has focused on the algorithmic issues, without investigating systems (OS and device) level aspects that affect IDD performance. 

Thus, despite the ubiquity of SSDs, some basic questions regarding IDD have not been investigated, including: What are reasonable baseline approaches to IDD, how do they perform, and where are the bottlenecks that need to be addressed by more sophisticated methods (such as compression)?
This paper seeks to fill this gap through a focused re-evaluation of IDD baselines. 

After a review of preliminaries and related work, %
we start %
by evaluating the behavior of the simplest possible baseline approach to IDD, which is to simply run \astar and rely on the OS virtual memory subsystem to use external memory when RAM is exhausted.
Then, we propose \siddastar, a simple, baseline algorithm for IDD-based external \astar search. 
Although previous work focused on open addressing implementations of hash tables for IDD, which necessitated methods such as compression, \siddastar uses a straightforward separate chaining based implementation without user-level caching, buffering, or multi-layer data structures.
We use \siddastar to investigate the effects of the page cache and to better understand  the system-level bottlenecks in IDD. 
We find that under high memory pressure, \siddastar can achieve node expansion rates comparable to A*-IDD, while achieving expansion rates significantly higher than A*-IDD and closer to RAM-based \astar in low memory-pressure situations.

\section{Preliminaries and Related Work}
\label{sec:preliminaries}

\subsection{SSD File Access in Linux}
\label{sec:linux-file-access}

I/O access in a standard OS such as Linux involves several layers of abstraction. In Linux, when a C++ program writes binary data to a file %
the request enters the kernel through a system call, where the Virtual File System routes it to the correct filesystem. The data is first copied into the kernel's page cache (in RAM), and the filesystem later translates the file offset into logical block addresses. The block layer sends these as I/O requests to the NVMe driver, which submits NVMe commands over PCIe to the SSD device. The SSD’s controller then maps the logical addresses to physical NAND flash cells %
and programs the data into flash memory. Each of these layers (page cache, block layer, NVMe SSD device) potentially incurs performance costs.

We consider two I/O interfaces: \texttt{pwrite}, which performs explicit system calls and copies data through the page cache, and \texttt{mmap}, which accesses files via memory mapping and page faults. The latter reduces syscall overhead but provides less explicit control over write timing.

\subsubsection{Page Cache}

The Linux page cache stores file data in RAM, allowing reads and writes to be served from memory. Writes are buffered and flushed asynchronously, and cached pages are evicted under memory pressure.

It is possible to bypass the page cache with
Direct I/O, a file access mode in which data transfers occur directly between user-space buffers and the storage device, bypassing the kernel’s page cache. In Linux, this behavior is typically enabled by opening a file with the \texttt{O\_DIRECT} flag.

\subsection{Previous work on external-memory \astar}

The primary bottleneck in external-memory search is duplicate detection, which determines whether a generated node is a duplicate of a previously generated node. This is a bottleneck because duplicate detection requires determining whether any previous record of the node exists in either RAM or external memory.

Previous work on external-memory search can be broadly classified based on how they perform duplicate detection,  
Most previous work focused on 
Delayed Duplicate Detection (DDD), an approach which does {\it not} perform duplicate detection on individual nodes when they are generated, and instead periodically performs a duplicate detection phase on large batches of nodes. For example, sorting-based DDD writes all newly generated nodes to a temporary file, sorts the file, and removes duplicates in a linear scan of the sorted file \cite{Korf:2003:DDD:1630659.1630926, edelkamp2004external}. 
Another approach is hash-based DDD, \cite{Korf:2004:BFS:1597148.1597253,DBLP:journals/jacm/Korf08, Korf:2016:CSA:3060621.3060707,DBLP:journals/jair/HatemBR18,DBLP:conf/ecai/SiagSFS24}. 

DDD has been shown to be highly effective in some domains, but DDD is not a simple modification of \astar. %
In DDD, node expansion order (among nodes with the same $f$-value) is strongly influenced by the specific DDD mechanism (e.g., the sort order in sorting-based DDD), making expansion order and  duplicate detection {\it non-orthogonal}. For example, it is not trivial to implement complex tie-breaking strategies among nodes with the same $f$ and $h$ values which have been shown to influence search efficiency in RAM-based \astar \cite{asai2017tie} with DDD. 

Immediate Duplicate Detection (IDD) is an approach where the expansion order  can, in principle, be exactly the same as standard RAM-based \astar, except that some/most of the data structures are stored on external memory instead of RAM. 
In {\it eager} IDD, the duplicate detection step (checking \closed) is performed immediately after node generation.
In {\it lazy} IDD, duplicate detection is performed before a node is expanded.
Eager IDD requires less RAM but incurs more duplicate detection checks, whereas lazy IDD uses more memory but incurs fewer duplicate detection checks.

In IDD, as with standard \astar, node expansion order and duplicate detection are much more orthogonal than in DDD, and 
IDD preserves the ability to implement the same best-first priority orders as \astar (including tie-breaking among nodes with the same $f$ and $h$ values). 
However, the large gap in random access latency between RAM and external memory can result in significantly slower node expansion rates compared to RAM-based \astar. 
Previous work proposed approaches to address this problem using in-memory (RAM) buffers and hash tables to minimize  external-memory accesses  \cite{EDELKAMP2011136,Edelkamp:2010:HST:1875144,DBLP:conf/aaai/LinF18}.

\subsubsection{Open addressing and Separate Chaining for \closed}

In standard \astar, \closed is typically one of the largest data structures and is represented as a hash table to support the query, "has state $s$ been seen before?"
Hash table implementations can be broadly categorized into approaches based on {\it separate chaining} or {\it open addressing}.
In separate chaining, each table index entry holds a chain of states that have the same hash value, and hash collisions are resolved by traversing the chain until an entry that matches the state is found.

In contrast, in open addressing, each table entry contains only one item. Collisions are resolved by {\it probing} additional slots in some sequence until a match is found.

Previous work on IDD has focused on open addressing.
Early theoretical work on IDD considered separate chaining as an option for representing  \closed, but dismissed it in favor of open addressing because chaining incurs storage overhead (an explicit pointer to the next element in the chain, which is unnecessary in open addressing), and processing overhead for reading the next element (compared  to open addressing with linear probing, which can read multiple entries in a single read I/O operation) \cite{Edelkamp:2010:HST:1875144,EDELKAMP2011136}.

In a basic implementation of \closed as an open addressed hash table on the SSD, each table entry contains all data for the node, and each insertion of a node into this table incurs a relatively expensive random write operation into some location in the file representing the table \cite{Edelkamp:2010:HST:1875144}, with little access locality.

\subsubsection{Compression and A*-IDD}

Previous work on IDD-based \astar is based on a technique called compression \cite{EDELKAMP2011136}, which separates a hash table into a RAM-resident portion and an external portion.

Compression represents \closed as follows.   
The (uncompressed) external table holds the actual node data (state values and  $f$, $g$ values) and is an array where entries are appended sequentially.
The (compressed) internal table is an open addressed hash table in RAM.
This design greatly alleviates  the random write bottleneck of the basic open addressing table design described above by always sequentially appending to the external table, as well as  reducing reads using the internal table.

Each entry in the internal table is a pointer to an entry in the external table on the SSD.
Given a  node $n$ for state $s$, duplicate detection in \closed works as follows. First, the internal table is probed for $s$ . If this probe  fails, then we know there is no duplicate in \closed (without having to access the external table).
If the probe returned a valid pointer, then we check external[p]. If external[p].state = s, then we found a duplicate. Otherwise, external[p].state was a hash collision, so we continue by probing the next entry in the internal table (according to the open addressing). This continues until either the probe fails ($n$ is not a duplicate), or we find a pointer to an external table entry whose state is equal to $s$ ($n$ is a duplicate).
In addition, a small buffer in RAM  %
can be used a cache, as well as a means to batch the writes to reduce the number of writes to SSD.

A*-IDD \cite{DBLP:conf/aaai/LinF18} introduced {\it segmented compression}, where  states are  mapped to a segment (among $p$ partitions) of the external table. False positive probes into the external table would require both a segment and entry collision, thus reducing the probability of false positives (and the accompanying expensive read access to SSD) by a factor of $p$. A*-IDD uses lazy duplicate detection.

\subsection{Experimental Setup and Preliminaries}

\subsubsection{Experimental Platform.}
All experiments were conducted on a machine equipped with an Intel Core i7-14700KF CPU, 32\,GiB of DDR5-5600 memory (2\,\(\times\)\,16\,GiB Micron DIMMs), and a PCIe Gen4 NVMe SSD (Samsung 990 series, manufacturer-rated 4\,KiB random IOPS: 1400K read and 1550K write).
Unless otherwise noted, experiments were performed on Ubuntu 24.04.3 LTS under identical hardware and software configurations.

\subsubsection{Memory Restriction and Measurement}
Memory is restricted using the cgroup v2 memory controller, \texttt{memory.max} setting which limits all RAM used by the process, including the page cache.
For memory usage, we report the peak memory consumption using the cgroup v2 \texttt{memory.peak} metric, which captures all memory charged to the cgroup during execution, including anonymous memory, file-backed page cache, and relevant kernel-side allocations.
To reduce inter-run interference, we drop the page cache before each measurement run.

\subsubsection{Baseline Search Algorithm and Heuristics}
We evaluate IDD-based \astar for heuristic-search based domain-independent, classical planning.
 As the RAM-based baseline, we use the Fast Downward (FD) implementation of \astar\ \cite{Helmert:2006:FDP:1622559.1622565}, which performs eager duplicate detection and uses a $(f,h,\mathrm{FIFO})$ tie-breaking policy.

We evaluate the search algorithms with two representative heuristics: \texttt{blind} and \texttt{merge-and-shrink} \cite{Helmert:2014:MAM:2628069.2559951}.
The \texttt{blind} heuristic has negligible computation cost, allowing the experiments to expose differences in external I/O performance more clearly.
Both heuristics are admissible in our unit-cost STRIPS setting; \texttt{merge-and-shrink} is also consistent and lookup-table based. For this study, we ignore the time cost of precomputing the heuristic tables, although under RAM-limited settings the RAM limit is increased by a domain-dependent amount ($\sim 100\,\mathrm{MiB}$) for precomputation.

\subsubsection{Benchmark Instances}
We used STRIPS planning tasks from the Autoscale benchmark suite \cite{Torralba_Seipp_Sievers_2021}, derived from IPC benchmarks.
Rather than evaluating all domains, we constructed three benchmark sets tailored to the goals of our experiments: a \texttt{blind} set, a \texttt{blind-easy} set, and a \texttt{merge-and-shrink} set.

For each set, we first formed a candidate pool among all domains from the Autoscale benchmarks for which RAM-based Fast Downward \astar\ solved at least one instance within a fixed time and memory budget, selecting at most one instance per domain.
We then chose smaller representative subsets for the reported experiments, favoring instances that were sufficiently search-intensive and informative for analyzing external-memory behavior under memory pressure, while removing redundant cases with highly similar runtime and memory profiles.

The resulting \texttt{blind} and \texttt{merge-and-shrink} benchmark sets consist of instances that already require several GiB of RAM for both standard RAM-based \astar and \siddastar.
The \texttt{blind-easy} set consists of lighter blind-heuristic instances and is used for the backing-storage experiments in Section~\ref{sec:os-overhead}.
Detailed thresholds, candidate-pool sizes, and complete instance lists are given in Appendix~\ref{supp:sec:instance-selection}.

\section{\astar: A Trivial Baseline for IDD}
\label{sec:astar-as-idd-baseline}

The simplest possible IDD-based baseline for \astar\ is to run the standard in-memory \astar\ algorithm under a strict RAM limit and rely on the operating system's virtual memory mechanism once that limit is reached.
When the search requires more memory than fits in RAM, the OS must continually evict and reload pages, effectively using external storage as backing memory.

\begin{figure}
    \centering
    \includegraphics[width=1.0\linewidth]{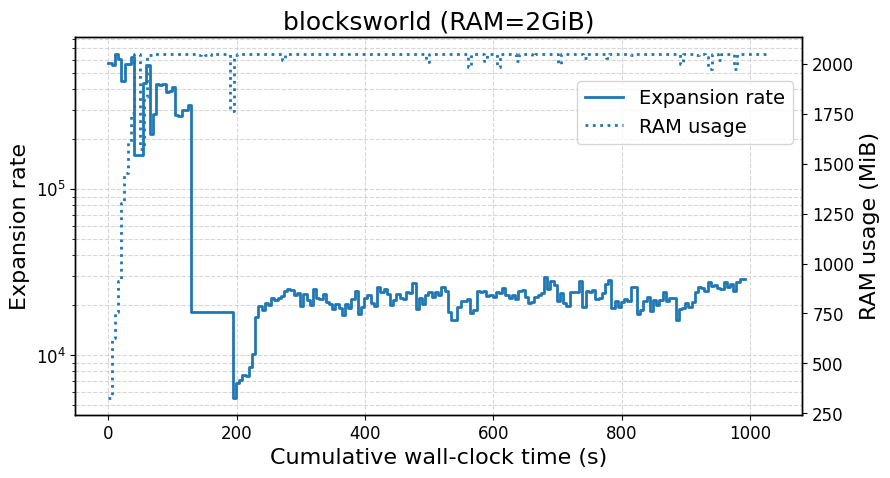}
    \caption{Expansion rate and RAM usage of Fast Downward \astar\ on \pddl{blocksworld-p23} with the \texttt{blind} heuristic under a 2\,GiB RAM limit. Expansion rate is based on nominal 5-second wall-clock sampling; during stall periods,  values may reflect expansions accumulated over longer intervals.}
    \label{fig:astar-swap}
\end{figure}

Figure~\ref{fig:astar-swap} shows the expansion rate of Fast Downward \cite{Helmert:2006:FDP:1622559.1622565} running \astar\ with the \texttt{blind} heuristic on \pddl{blocksworld-p23} under a RAM limit of 2\,GiB.
At the beginning of the search, the expansion rate is high, at roughly several $10^5$ nodes per second.
RAM usage reaches the 2\,GiB limit within the first few dozen seconds, but the sharp slowdown occurs only after a lag of roughly 100 seconds.
Around 200 seconds, the search undergoes a brief near-stall, then recovers to a much lower steady-state regime of roughly several $10^4$ nodes per second.
For the remainder of the run, RAM usage stays near the imposed limit while the search remains in this degraded low-throughput state.

This behavior is consistent with swap thrashing under severe memory pressure.
Once the working set no longer fits in RAM, accesses to \open and \closed trigger frequent page eviction and reloading, so the search spends much of its time waiting for memory transfers rather than expanding nodes.
Similar qualitative behavior was observed across other PDDL domains under RAM pressure; see Section~\ref{sec:memory-pressure} and Appendix  Figures~\ref{supp:fig:raw-blind-ram-limit} and \ref{supp:fig:raw-merge-and-shrink-ram-limit}.

\begin{observation}
This trivial approach is practical if the amount of RAM available is almost enough to solve the problem, but the orders of magnitude slowdown after RAM is exhausted leaves much room for improvement by more sophisticated IDD-based methods for problems that require much more memory than the RAM available. 
\end{observation}

\section{\siddastar}

\begin{algorithm}[t]
\caption{\siddastar}
\label{alg:sidd}
\small
\textbf{(a) Main loop}
\begin{algorithmic}[1]
\State $n_0 \gets (s_0, 0, h_0, \bot)$
\State $\textsc{OpenInsert}(n_0, \textsc{Key}(n_0))$
\While{$\textsc{Open} \neq \emptyset$}
    \State $n_u \gets \textsc{OpenExtractMin}()$
    \If{$\textsc{ClosedFindOrInsert}(n_u)$}
        \If{$\textsc{IsGoal}(s(n_u))$}
            \State \textbf{return} $\textsc{ExtractPlan}(n_u)$
        \EndIf
        \ForAll{$(\text{op}, s') \in \textsc{Successors}(s(n_u))$}
            \State $g' \gets g(n_u) + \textsc{Cost}(\text{op})$
            \State $h' \gets h(s')$
            \State $n_v \gets (s', g', h', n_u, \text{op})$
            \State $\textsc{OpenInsert}(n_v, \textsc{Key}(n_v))$
        \EndFor
    \EndIf
\EndWhile
\State \textbf{return} failure
\end{algorithmic}

\vspace{0.5em}
\textbf{(b) \textsc{ClosedFindOrInsert}}
\begin{algorithmic}[1]
\Procedure{ClosedFindOrInsert}{$n$}
    \State $b \gets \textsc{Bucket}(\textsc{Hash}(s(n)))$
    \State $i \gets \textsc{Head}[b]$ \Comment{RAM-resident bucket head}
    \While{$i \neq \textsc{Null}$}
        \State $(x, i_{\mathrm{next}}) \gets \textsc{ReadRecord}(i)$
        \If{$s(x)=s(n)$}
            \If{$g(n) < g(x)$}
                \State $\textsc{UpdateHeader}(i, n)$
                \State $\textsc{OpenInsert}(n, \textsc{Key}(n))$
            \EndIf
            \State \textbf{return} false
        \EndIf
        \State $i \gets i_{\mathrm{next}}$
    \EndWhile
    \State $u \gets \textsc{AppendNode}(n, \textsc{Head}[b])$
    \State $\textsc{Head}[b] \gets u$
    \State \textbf{return} true
\EndProcedure
\end{algorithmic}
\end{algorithm}

This section presents \emph{Simple Immediate Duplicate Detection \astar} (\siddastar), an external-memory implementation of \astar with lazy duplicate detection, 
which places \open and \closed on external storage.
\siddastar is intended as a minimal %
baseline for IDD-style external-memory heuristic search, and 
exposes the core interaction between best-first search and external storage through simple external-memory data structures.

A key design choice in \siddastar is the use of \emph{separate chaining} for \closed.
As discussed in the preliminary section, prior work on IDD-based external-memory \astar has primarily focused on open addressing \cite{Edelkamp:2010:HST:1875144,EDELKAMP2011136,DBLP:conf/aaai/LinF18}.
In contrast, \siddastar keeps a compact RAM-resident head table and stores collided records as linked chains on external storage.
The resulting implementation is simple and yields a different external-memory I/O pattern from open addressing-based designs, as discussed below in Section \ref{sec:separate-chaining-vs-open-addressing}.

Algorithm~\ref{alg:sidd} summarizes the main search loop and the \closed operation.
Nodes are extracted from \open, and duplicate detection is performed only when a node is removed from \open.
In Algorithm~\ref{alg:sidd}(b), \textsc{ClosedFindOrInsert} returns \textbf{true} iff $n$ is the first extracted node for its state to be inserted into \closed; in that case the node is expanded.
If the state is already present, the procedure returns \textbf{false}; if $n$ improves the stored $g$-value for that state, the procedure also updates the corresponding header and reinserts $n$ into \open.
Under nonnegative operator costs and a consistent heuristic, \siddastar returns an optimal solution and expands the same states as standard \astar modulo tie-breaking and duplicate-handling timing.
The main novelty is therefore in the external-memory realization of \open and especially \closed, rather than in the high-level control flow.

\subsection{\open Implementation}

\open is implemented as a simple external-memory two-level bucket queue.
For a node $n$, we define $\textsc{Key}(n)=(f(n),h(n))$, where $f(n)=g(n)+h(n)$.
Nodes are ordered by increasing $f$, ties are broken by increasing $h$, and nodes with equal $(f,h)$ are extracted in FIFO order.

Because duplicate detection is deferred until node extraction, \open stores complete search nodes rather than node identifiers.
Each node record contains the packed state representation together with its parent pointer, generating operator, and $g$- and $h$-values.

For each $(f,h)$ pair, \siddastar maintains a separate bucket file on external storage.
Insertion appends a node record to the file corresponding to its $(f,h)$ value, and \textsc{OpenExtractMin} removes a node from the lexicographically smallest non-empty $(f,h)$ bucket in FIFO order.

Because multiple \open entries for the same state may coexist, duplicate entries may remain in \open until extraction.
Such entries are handled by \textsc{ClosedFindOrInsert}.

Only lightweight metadata is kept in RAM, namely the set of non-empty $f$-layers and, within each layer, the set of non-empty $h$-buckets. In the implementation used here, bucket files are managed without user-level buffering.

\subsection{\closed with External Separate Chaining}
\closed supports three operations: exact duplicate detection, insertion of newly extracted states, and access to per-state metadata, including the parent pointer, generating operator, and the stored $g$- and $h$-values.

A key design choice of \siddastar to implement \closed as an external-memory hash table with separate chaining.
Only the bucket-head table is kept in RAM, requiring just one 4-byte pointer per bucket; the collision chains themselves are stored externally.
Each \closed record consists of an index part (hash value and next pointer), a fixed-size node header, and the packed state representation.
The node header stores the parent pointer, generating operator, and the $g$- and $h$-values associated with the current best path recorded for that state.

Given a state $s$, \siddastar computes a 32-bit hash value, maps it to a bucket in $\{0,\dots,B-1\}$,
where $B$ is the number of buckets in the head table, 
and follows the linked chain starting from the RAM-resident head pointer.
For each visited record, the packed state is read from external storage and compared exactly against $s$.
If a matching state is found, the stored header provides the current metadata for that state.
When the new path is cheaper, \siddastar updates only the header of the existing record and reinserts the improved node into \open.
If no match is found, a new record is appended to the external record file and linked at the head of the corresponding chain.

As in \open, the implementation used here does not employ user-level buffering for \closed.
New states are appended to an external record file, so the dominant \closed write pattern is append-only, except for occasional header updates for improved duplicates.
Separate chaining therefore keeps the bucket directory compact and RAM-resident, while making duplicate detection depend on chain traversal and external record reads.

\subsection{\siddastarpwrite and \siddastarmmap (\texttt{pwrite} vs. \texttt{mmap})}
Although the algorithmic behavior of \siddastar is defined above, performance also depends on the I/O interface used to access the external \open and \closed files.
These files can be accessed either via explicit file I/O using \texttt{pread}/\texttt{pwrite} or via a memory-mapped interface using \texttt{mmap}.
We therefore consider two implementations:
\siddastarpwrite, which uses \texttt{pread}/\texttt{pwrite}, and
\siddastarmmap, which uses \texttt{mmap}.
Both implementations use the same search algorithm and data structures, differing only in the mechanism used to access external storage.
We do not use hinting interfaces such as \texttt{posix\_fadvise} or \texttt{madvise}.
Experiments with \texttt{O\_DIRECT} necessarily use \texttt{pread}/\texttt{pwrite}, since \texttt{mmap} operates through the page cache.
Below, we use \siddastar when the text applies to both \siddastarpwrite and \siddastarmmap.

\subsection{\siddastar/OA: A Baseline Implementation with Open Addressing}
\label{sec:sidd-open-addressing}

To better understand the design tradeoffs in external-memory duplicate detection, we also implemented a baseline variant of \siddastar in which the external \closed is realized using \emph{open addressing} rather than separate chaining.
This variant, \siddastar/OA, uses the same search algorithm and external-memory \open as \siddastar, and differs only in the representation of \closed.

In \siddastar/OA, \closed is a fixed-size external hash table with $M$ slots, no deletion, and no resizing during search.
Collisions are resolved by linear probing.
Each slot stores one complete \closed record: an index part (occupancy flag and 32-bit hash), a fixed-size node header, and the packed state representation.
Lookup reads the index part of each probed slot and reads the full record only when the stored hash matches; insertion writes the complete record to the first empty slot in the probe sequence, while improved duplicates update only the header.
The implementation uses no RAM-resident auxiliary index and no user-level buffering or batched writes.
Additional implementation details and load-factor statistics are given in Appendix~\ref{supp:sec:oa-details}.

\subsection{Separate Chaining vs. Open Addressing}
\label{sec:separate-chaining-vs-open-addressing}
In our baseline implementations, the most important difference between separate chaining and open addressing is the write path for newly inserted \closed records.
Both implementations perform file-backed reads during duplicate detection, so probing itself is not the main distinction in our comparison.
With separate chaining, each new record is written by appending it to the end of the external record file, while the bucket-head table is updated only in RAM.
This yields sequential full-record writes, except for occasional small random header updates for improved duplicates.
With open addressing, by contrast, insertion writes the complete record directly into the first empty slot found along the probe sequence.
Because the destination slot depends on the hash value and collision pattern, this results in random full-record writes.
We therefore expect the dominant I/O disadvantage of straightforward open addressing to come primarily from record insertion, rather than from probing alone.
Sequentially appending such records is generally favorable for file-system and storage-layer behavior, whereas random placement of full records is more likely to incur higher overhead, especially when memory pressure is high and the page cache cannot absorb most accesses.

\subsection{\siddastar-Based Classical Planner}

We implemented a domain-independent planner based on \siddastar.
The implementation is derived from Fast Downward (FD)\cite{Helmert:2006:FDP:1622559.1622565}, replacing the standard \astar component with \siddastar. Much of the FD code is reused, including successor generation, task representation, and %
heuristic functions. 
States use the same packed encoding and 32-bit hash computation as FD, ensuring identical state-to-key mappings.
The differences lie in the implementation of the underlying data structures for \open and \closed.

\section{Experimental Evaluation of \siddastar}

We evaluate and compare:
(1) the standard, RAM-based implementation of \astar in Fast Downward, with virtual memory page-swapping enabled, 
(2) A*-IDD, 
(3) \siddastarpwrite,
(4) \siddastarmmap,
(5) \siddastarpwrite/OA
(6) \siddastarmmap/OA

The A*-IDD implementation is based on the original implementation by Lin (\citeyear{lin-a*-idd-code}).
Although the original implementation uses LIFO tie-breaking, we modified A*-IDD to use FIFO tie-breaking to match the tie-breaking policies of  \astar and \siddastar.
We verified that A*-IDD search performance with FIFO tie-breaking was very similar to search performance with LIFO tie-breaking (Appendix  \ref{supp:sec:a*-idd-fifo-vs-lifo-tiebreak}).

Unless otherwise noted, both the RAM-resident head table of \siddastar (separate chaining) and the internal table of A*-IDD were set to 512\,MiB.
For \siddastar/OA, the external hash table size was fixed to $2^{27}$ slots, matching the bucket count of the separate chaining variant.

\ifdefined\verifiedexpansions
We verified that when executed to completion without RAM or runtime limits, all the algorithms expanded the same number of nodes on all benchmark instances used in this paper (see Appendix \ref{supp:sec:eager-vs-lazy-dd}).
Since all implementations expanded the same number of nodes in unrestricted runs, expansion rate is an appropriate implementation-level comparison metric for this study.

\else
All of these algorithms  expand the same number of nodes on all problems that were solved. 
Furthermore, Fast Downward's standard  \astar implementation (eager duplicate detection) and a modified version with lazy duplicate detection expand the same number of nodes on all benchmark instances (Appendix \ref{tab:eager-vs-lazy-expanded}), supporting the use of expansion rate as the primary comparison metric.

Since the algorithms have essentially the same search behavior, 
the main performance metric to evaluate IDD implementations is the node expansion rate (expansions per wall-clock second), which depends on the data structures and implementation details compared in this paper.
\fi

To reduce variability caused by file growth during search, \siddastar preallocates its external files at the beginning of the first run using \texttt{fallocate} (details in Appendix \ref{supp:sec:preallocation}).

Additional details on experimental settings are in the Appendix \ref{supp:sec:experiments-additional-details}. %

\begin{table*}[htb]
\centering
\small
\setlength{\tabcolsep}{4pt}
\renewcommand{\arraystretch}{1.1}

\begin{tabular}{l|rr  G rr  Grr|rr}
\hline
Problem & \multicolumn{6}{c|}{{\bf 1.0GiB RAM limit}} & \multicolumn{2}{c}{{\bf 600MiB RAM limit}} \\

& \astar
& A*-IDD
& \siddastarpwrite
& \siddastarmmap
& \siddastarpwrite
& \siddastarmmap
& A*-IDD
& \siddastarpwrite \\
&
&

& Sep. Chaining
& Sep. Chaining
& Open Address.
& Open Address.
& 
& Sep. Chaining\\
\hline
\multicolumn{9}{l}{{\bf Blind heuristic}}\\
\hline
\pddl{blocksworld-p23} & 29{,}181 & \textbf{70{,}565} & 29{,}771 & 19{,}454 & 9{,}946 & 2{,}383 & {\bf 45{,}051} & 28{,}546 \\
\pddl{data-network-p08} & 3{,}471 & 19{,}056 & \textbf{30{,}667}$^{*}$ & 11{,}976$^{*}$ & 4{,}480 & 309 & {\bf 12{,}108} & 11{,}553 \\
\pddl{depots-p23} & 6{,}675 & 16{,}789 & \textbf{27{,}586} & 16{,}429 & 3{,}424 & 1{,}000 & 6{,}565 & {\bf 18{,}842} \\
\pddl{floortile-p03} & 3{,}543 & \textbf{44{,}457} & 43{,}042 & 23{,}561 & 5{,}975 & 231 & 19{,}070 & {\bf 27{,}832} \\
\pddl{mprime-p05} & 8{,}238$^{*}$ & 25{,}194 & \textbf{49{,}111}$^{*}$ & 29{,}558$^{*}$ & 4{,}716 & 1{,}513 & 11{,}746 & {\bf 20{,}880} \\
\pddl{rovers-p03} & 27{,}881 & 28{,}693 & \textbf{44{,}050} & 20{,}353 & 4{,}655 & 1{,}346 & 14{,}951 & {\bf 32{,}717} \\
\pddl{snake-p22} & 26{,}564 & \textbf{48{,}140} & 30{,}560 & 22{,}658 & 7{,}381 & 2{,}412 & 29{,}104 & {\bf 30{,}837} \\
\pddl{storage-p05} & 25{,}231 & \textbf{63{,}121} & 30{,}192 & 19{,}660 & 6{,}208 & 2{,}825 & {\bf 31{,}781} & 24{,}419 \\

\hline
\multicolumn{9}{l}{{\bf Merge-and-shrink heuristic}}\\

\hline
\pddl{agricola-p09}     & \textbf{72{,}137}$^{*}$ & 30{,}015 & 32{,}856 & 27{,}451 & 11{,}167 & 2{,}933 & 28{,}003 & {\bf 41{,}019} \\
\pddl{blocksworld-p10}  & 36{,}069$^{*}$ & 51{,}131$^{*}$ & \textbf{158{,}287}$^{*}$ & 101{,}682$^{*}$ & 6{,}857 & 2{,}519 & 40{,}398$^{*}$ & {\bf 72{,}227$^{*}$} \\
\pddl{data-network-p18} & 5{,}371 & 11{,}120 & \textbf{28{,}937} & 23{,}223 & 1{,}347 & 685 & 5{,}435 & {\bf 21{,}536} \\
\pddl{depots-p05}       & 15{,}083 & 14{,}422 & \textbf{29{,}042} & 11{,}777 & 4{,}059 & 1{,}877 & 12{,}730 & {\bf 24{,}327} \\
\pddl{driverlog-p26}    & 4{,}436 & \textbf{33{,}607}$^{*}$ & 104{,}731$^{*}$ & 89{,}403$^{*}$ & 5{,}131 & 1{,}852 & 19{,}572 & {\bf 48{,}477$^{*}$} \\
\pddl{floortile-p07}    & 3{,}855 & \textbf{37{,}331} & 25{,}357 & 21{,}510 & 4{,}926 & 1{,}256 & 17{,}295 & {\bf 21{,}769} \\
\pddl{hiking-p17}       & \textbf{66{,}122} & 25{,}892 & 50{,}747 & 27{,}486 & 5{,}251 & 2{,}136 & 24{,}384 & {\bf 34{,}711} \\
\pddl{zenotravel-p08}   & 8{,}119 & 17{,}797 & 58{,}568$^{*}$ & \textbf{63{,}958}$^{*}$ & 3{,}691 & 2{,}008 & 13{,}891 & {\bf 31{,}548$^{*}$} \\
\hline
\end{tabular}

\caption{Exp. rates (states/sec) for the last 60 seconds of search, under high memory pressure. 
A ``$^{*}$'' indicates search completed.}
\label{tab:combined-results}
\end{table*}

\subsection{Comparison Under High Memory Pressure}
\label{sec:comparison-under-high-memory-pressure}

We compared the algorithms on 8 instances each for \texttt{blind} and \texttt{merge-and-shrink} heuristics. 
All 6 algorithms were compared with a 1GiB RAM limit. 
This is a low RAM limit compared to the amount of storage required to solve these problem instances, i.e., memory pressure is high.
Each run was executed with a 600 second time limit.

In addition, the two algorithms which performed best under 1GiB ( \siddastarpwrite and A*-IDD) were  compared under even higher memory pressure (600MiB RAM limit).

To focus on the expansion rates when the memory pressure is highest, Table \ref{tab:combined-results} shows the number of node expansions per second on the last 60 seconds of the run (the expansion rate over the entire run are in the Appendix Table \ref{supp:tab:combined-full-expansion-rate}). Runs that solved the instance are marked with a $*$.

\begin{observation}
The straightforward open addressing baseline performs extremely poorly for IDD-based \astar.     
\end{observation}

Appendix~\ref{supp:tab:oa-load-factor} shows that the OA runs had low load factors at the 600-second cutoff (below 0.05 in all reported RAM-limited runs), so this poor performance is not simply due to near-full-table behavior.

\begin{observation}
Straightforward separate chaining (\siddastar) performs much better, and is comparable to the state-of-the-art approach (A*-IDD) on these benchmarks. 
\end{observation}

\subsection{On the Importance of the Page Cache}

The importance of the page cache in the context of external-memory search was alluded to by \citeauthor{DBLP:conf/aaai/LinF18} (\citeyear{DBLP:conf/aaai/LinF18}), but not investigated.
In general, more RAM available for the page cache will result in better I/O latency.
Thus, there is a tradeoff between the amount of RAM used by data structures for improving performance in IDD-based search vs. the amount of RAM left available for the page cache.

\begin{table}[htb]
\centering
\scriptsize
\setlength{\tabcolsep}{3pt}
\renewcommand{\arraystretch}{1.1}
\resizebox{\columnwidth}{!}{%
\begin{tabular}{l|rr|rrGrr}
\hline
Problem
& \multicolumn{2}{c|}{\astar}
& \multicolumn{2}{cG}{\siddastarmmap}
& \multicolumn{2}{c}{\siddastarpwrite} \\
& Exp. rate & RAM & Exp. rate & RAM & Exp. rate & RAM \\
\hline
\pddl{blocksworld-p23}  & 613{,}812 & \textbf{5{,}952} & \textbf{619{,}260} & 8{,}842 & 275{,}866 & 8{,}829 \\
\pddl{data-network-p08} & \textbf{209{,}699} & \textbf{4{,}347} & 127{,}829 & 7{,}483 & 61{,}915 & 7{,}466 \\
\pddl{depots-p23}       & \textbf{410{,}784} & \textbf{3{,}214} & 226{,}325 & 8{,}993 & 95{,}376 & 8{,}977 \\
\pddl{floortile-p03}    & \textbf{442{,}277} & \textbf{3{,}859} & 386{,}156 & 7{,}646 & 168{,}778 & 7{,}627 \\
\pddl{mprime-p05}       & \textbf{331{,}365} & \textbf{3{,}077} & 235{,}846 & 9{,}197 & 101{,}557 & 9{,}181 \\
\pddl{rovers-p03}       & \textbf{599{,}220} & \textbf{3{,}420} & 228{,}980 & 6{,}289 & 83{,}819 & 6{,}273 \\
\pddl{snake-p22}        & \textbf{337{,}423} & \textbf{5{,}524} & 319{,}380 & 6{,}687 & 235{,}724 & 6{,}671 \\
\pddl{storage-p05}      & \textbf{664{,}610} & \textbf{3{,}249} & 512{,}314 & 7{,}346 & 217{,}441 & 7{,}329 \\
\hline
\end{tabular}%
}
\caption{Expansion rates (states/sec) and peak RAM usage (cgroup, MiB)  on 8  instances (blind heuristic) with no RAM limit (i.e., full usage of 32GB system RAM).} %
\label{tab:expansion-rate-ram-by-method}
\end{table}
\subsubsection{Evaluation of \siddastar with no RAM limit}

We evaluated \astar and \siddastar on 8 instances using the \texttt{blind}  heuristic. 
Each run was executed without a  RAM limit (i.e., full usage of the 32GB RAM), and all runs were executed until completion. 
All of these problem instances are solvable by \astar in the available RAM, so this is a low memory pressure scenario.
Data using the \texttt{merge-and-shrink} heuristic is in the Appendix Table \ref{tab:expansion-rate-ram-by-method-mands}.

Table \ref{tab:expansion-rate-ram-by-method} shows expansions/second and the RAM (including page cache) used by each method.
These results show that with sufficient RAM for the page cache such that the entire search can fit in the page cache, \siddastar can expand nodes at rates which are comparable to \astar (within a factor of 2 for \siddastarmmap, and within a factor of 8 for \siddastarpwrite).

\begin{observation} 
Under low memory pressure, the page cache significantly boosts IDD  performance; if the RAM available is sufficient for the entire search,  expansion rates can be within an order of magnitude of  RAM-based \astar.
\end{observation}

\subsubsection{A*-IDD internal table size vs page cache}
In \cite{DBLP:conf/aaai/LinF18}, the authors sidestepped the interaction between page cache and A*-IDD 
by using almost all of the available RAM for the internal hash table for compression (they state that they chose the internal table size "to limit
any speedups from page caching while provisioning enough
space for Fast Downward" (Sec. 5.3, par.3)).

Table \ref{tab:a*-idd-internal-table-experiment} compares A*-IDD with an internal table size of 900MiB vs A*-IDD with an internal table size of 512MiB, both running under a 1GiB RAM limit. The version using the {\it smaller} internal table has a significantly higher expansion rate than the one with the larger table.

\begin{observation}
IDD algorithm data structures compete with the page cache for RAM. Using {\it less} RAM for the IDD-related data structures leaves more RAM for the page cache, which can sometimes {\it improve} performance.
\end{observation}

\begin{table}[htb]
\centering
\small
\setlength{\tabcolsep}{4pt}
\renewcommand{\arraystretch}{1.1}
\begin{tabular}{l|rrrrrr}
\hline
Problem
& A*-IDD (512MiB)
& A*-IDD (900MiB)\\
\hline
\multicolumn{3}{l}{{\bf Blind heuristic}}\\
\hline
\pddl{blocksworld-p23} & 78,945 & 45,957 \\
\pddl{data-network-p08} & 13,742 & 6,134 \\
\pddl{floortile-p03} & 24,173 & 10,129\\
\hline
\multicolumn{3}{l}{{\bf Merge-and-shrink heuristic}}\\
\hline
\pddl{agricola-p09}     & 20,764 & 20,018  \\
\pddl{data-network-p18} & 9,359 & 6,710\\
\pddl{floortile-p07}    & 34,403 & 13,483\\
\hline
\end{tabular}
\caption{A*-IDD comparison of internal table sizes (900MiB vs 512MiB) with 1GiB RAM limit}
\label{tab:a*-idd-internal-table-experiment}
\end{table}

\subsection{Where are the overheads for external-memory access?}
\label{sec:os-overhead}

In this section, we try to quantify how much of the I/O overheads is due to the SSD device, and how much of the overhead is due to operating-system level overheads.

In order to measure system-related overheads independent of the external-memory device, we use {\it memory-backed filesystem}, where a portion of  system RAM is made accessible using the file system interface.
A \brd (block ram disk) is a block-layer abstraction where a fixed size block of RAM is made available as a block device (e.g., /dev/ram0).  %
Another memory-backed filesystem is \tmpfs, a filesystem-layer abstraction which implements a dynamically sized, memory-backed filesystem. %
These memory-backed filesystems can be used instead of an SSD to store the external-memory data structure, and can be used as a proxy for an ideal SSD with no device-level overheads.

This allows us to compare 3 types of storage for the external data structures, each of which incurs different overheads: (1) \ssd:  device overhead + block driver overhead + filesystem overhead, (2) \brd: block driver  + filesystem , and (3) \tmpfs: filesystem.
By comparing the performance of \siddastar using \ssd, \brd, and \tmpfs, we can better understand the relative contribution of each of the 3 layers of overhead (device, block driver, filesystem) in this setup.

\subsubsection{With Page Cache - No memory pressure}

Table \ref{tab:overhead-layers} compares 
the node expansion rates of \ssd, \brd, and \tmpfs 
in a low memory pressure scenario (RAM limit=32GB) using instances which can all be solved using less than  6GB of RAM by \astar.
Although \tmpfs has the highest expansion rate by a very small margin, \brd and \ssd have almost the same expansion rates.
Despite the fact that \ssd, \brd, and \tmpfs incur different levels of overhead, their performance is almost indistinguishable in this scenario.

\subsubsection{Without Page Cache (\texttt{O\_DIRECT})}

Next, we configure the I/O to bypass the page cache and use Direct I/O by using the \texttt{O\_DIRECT} flag when opening the files on the storage layer.
Table \ref{tab:overhead-layers} shows that without a page cache to mask the overheads, \tmpfs is 2-3 times faster than \brd, which in turn is almost 2 orders of magnitude faster than \ssd.

\begin{observation} 
The device, block driver, and file system all incur significant overheads. However, in low-memory pressure conditions, the page cache can hide the block layer and device overheads.
\end{observation}

\begin{table*}[htb]
\small
\centering
\begin{tabular}{l|r|rrr|rrr}
\hline
 & \astar (blind) & \multicolumn{3}{c}{\siddastarpwrite {\bf with} page cache}  & \multicolumn{3}{c}{ \siddastarpwrite {\bf without} page cache } \\
 &  & \tmpfs & \brd & \ssd & \tmpfs & \brd & \ssd \\
\hline
\pddl{barman-p02}        & 880,527 & 220,950 & 206,096 & 246,549 & 162,533 & 69,627 & 1,433 \\
\pddl{blocksworld-p04}  & 1,009,026 & 361,513 & 340,223 & 337,634 & 224,297 & 76,626 & 1,944 \\
\pddl{gripper-p06}      & 1,296,619 & 279,892 & 262,435 & 263,088 & 171,587 & 68,875 & 1,429 \\
\pddl{pegsol-p23}       & 1,254,337 & 423,892 & 462,881 & 460,173 & 307,257 & 129,459 & 2,767 \\
\pddl{termes-p01}       & 998,678 & 298,692 & 321,677 & 331,494 & 216,510 & 88,257 & 1,950 \\
\hline
Geom. mean (5) & 1,076,117 & 309,298 & 307,290 & 319,720 & 210,788 & 84,066 & 1,847 \\
\hline
\end{tabular}
\caption{Comparison of backing storage (\tmpfs, \brd, \ssd) with and without page cache (expansions/second)}
\label{tab:overhead-layers}
\end{table*}

\begin{figure*}[!htb]
  \centering

  \begin{subfigure}{0.33\textwidth}
    \centering
    \includegraphics[width=\linewidth]{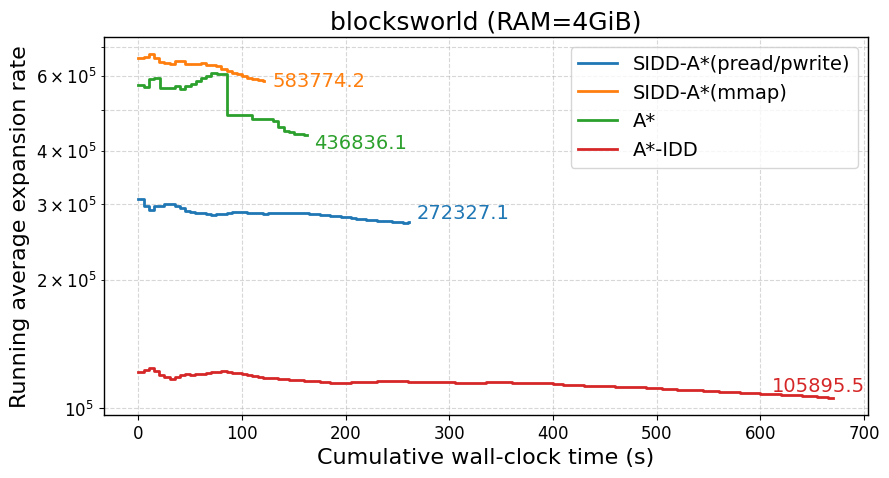}
  \end{subfigure}
  \hfill
  \begin{subfigure}{0.33\textwidth}
    \centering
    \includegraphics[width=\linewidth]{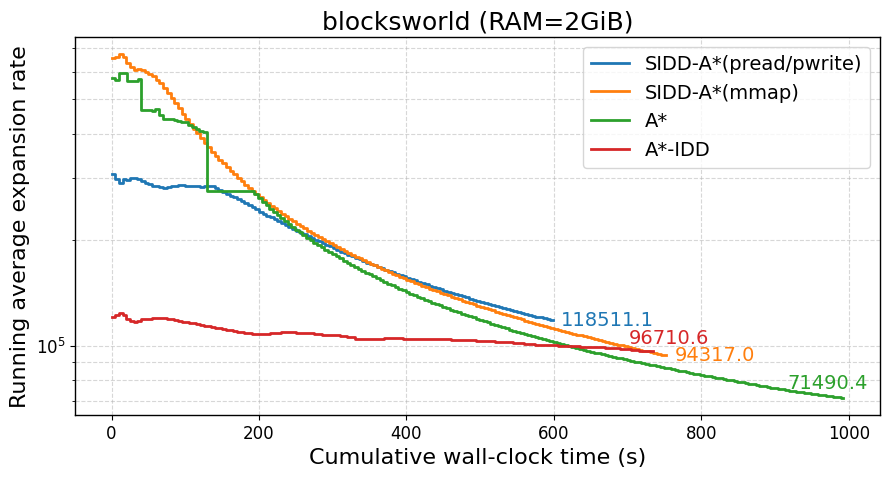}
  \end{subfigure}
  \hfill
  \begin{subfigure}{0.33\textwidth}
    \centering
    \includegraphics[width=\linewidth]{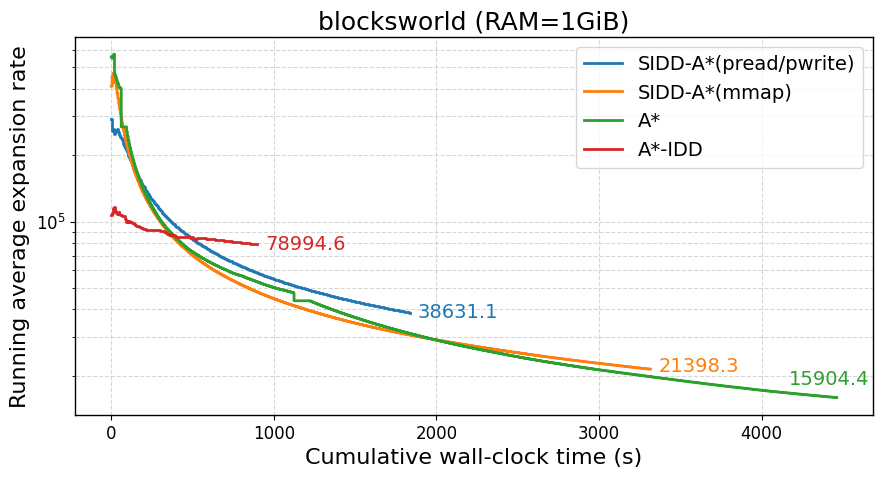}
  \end{subfigure}

  \begin{subfigure}{0.33\textwidth}
    \centering
    \includegraphics[width=\linewidth]{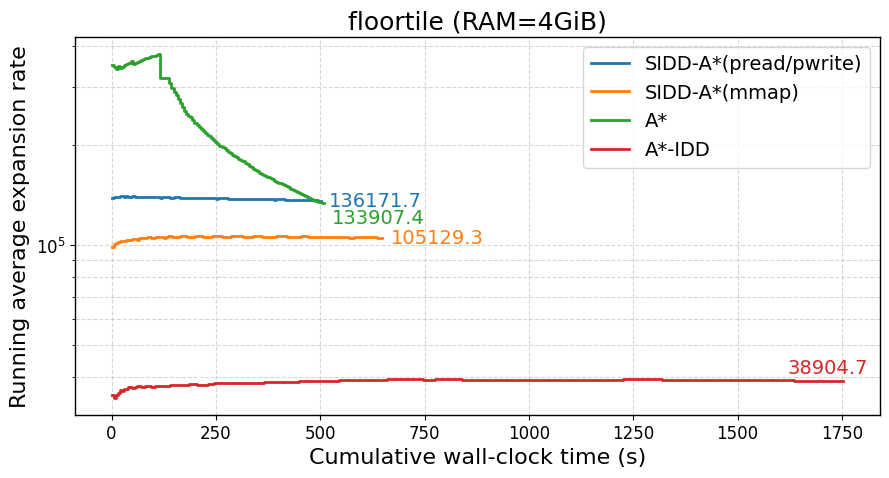}
  \end{subfigure}
  \hfill
  \begin{subfigure}{0.33\textwidth}
    \centering
    \includegraphics[width=\linewidth]{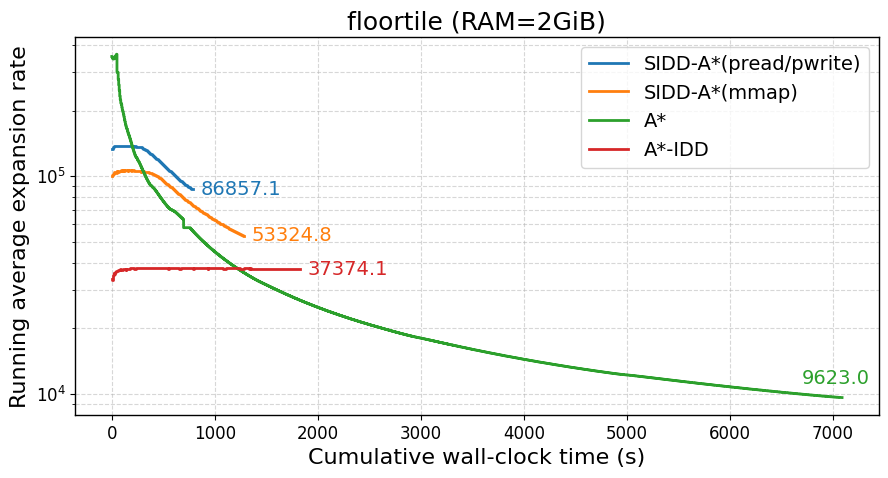}
  \end{subfigure}
  \hfill
  \begin{subfigure}{0.33\textwidth}
    \centering
    \includegraphics[width=\linewidth]{figures/floortile_1GIB_mands.png}
  \end{subfigure}
  
  \caption{running average expansion rate vs. wall-clock time for 4GiB, 2GiB, and 1GiB RAM limitation on \pddl{blocksworld-p23} (blind heuristic) and \pddl{floortile-p07} (merge-and-shrink heuristic)}
  \label{fig:cumulative-expansion-rate-vs-time}
\end{figure*}

\subsection{More Detailed View of Expansion Rate Behavior vs. Time}
\label{sec:memory-pressure}

To better understand the behavior of IDD-based \astar, we look at how the expansion rates change over time.
We also evaluate this behavior with 3 different RAM limits (4GiB, 2GiB, and 1GiB), to observe how the amount of system RAM available affects the expansion rate behavior.

Figure~\ref{fig:cumulative-expansion-rate-vs-time} plots
running average expansion rate vs. wall-clock runtime for \pddl{blocksworld-p23} (blind heuristic) and \pddl{floortile-p07} (merge-and-shrink heuristic), with 4GiB, 2GiB, and 1GiB RAM limits.
All runs were executed to completion (the end of each line denotes the time when that algorithm completed its search).
Running average expansion rate (total number of expansions up to time $t$ divided by $t$) is shown instead of the instantaneous expansion rate at time $t$ because the instantaneous expansion rates fluctuate and are difficult to see, especially with multiple plots per figure.

For all of the algorithms, the expansion rates decrease over time, and also  as the RAM limit is decreased.
\astar initially has a high expansion rate, but as RAM is depleted and exhausted, its expansion rate drops precipitously.
\siddastarpwrite and \siddastarmmap also have a relatively high initial expansion rate which drops over time.
A*-IDD expansion rates are relatively stable compared to \astar and \siddastar.

As the RAM limit is decreased, all algorithms exhibit degraded performance curves.
\astar slows down as the RAM limit decreases because reducing RAM increases page swap thrashing. The other algorithms slow down because of increased page cache misses/eviction. Compared to \siddastarpwrite, \siddastarmmap tends to degrade more both over time and also as the RAM limit is decreased.

Additional running-average, raw-rate, and total-expansion plots are in the Appendix (Figures~\ref{supp:fig:blind-ram-limit}--\ref{supp:fig:total-merge-and-shrink-ram-limit}).

\begin{observation} 
The performance of all evaluated algorithms depends on available RAM; even algorithms such as \siddastar that use very little RAM explicitly run faster  with additional RAM for the page cache.
\end{observation}

\begin{observation}
    \texttt{mmap}-based external data structure access can be more susceptible to degradation under RAM pressure than \texttt{pwrite}-based access. %
\end{observation}

\section{Conclusion and Discussion}

This paper experimentally evaluated baseline approaches for immediate duplicate detection in external-memory search.
We proposed \siddastar, 
which uses a simple \closed implemented with a straightforward separate chaining hash table, and showed that this is sufficient to achieve expansion rates comparable to the previous state-of-the-art algorithm, A*-IDD, without user-level caches or buffers. Building upon this simple baseline with more complex techniques to improve performance is a promising direction for future work.

We also investigated the performance impact of the OS page cache, and showed that in low memory pressure situations, a sufficient amount of page cache can significantly speed up IDD. We also showed that significant I/O overheads can be attributed not only to the SSD device, but also the OS block device driver and filesystem system call layers.  
Therefore, approaches that bypass the OS layers in order to better control performance at the application level are a direction for future work.

\FloatBarrier
\bibliography{ems}

\ifdefined\arxiv %
\clearpage
\appendix

\setcounter{secnumdepth}{3} %
\setcounter{table}{0}
\setcounter{figure}{0}
\renewcommand{\thefigure}{A\arabic{figure}}
\renewcommand{\thetable}{A\arabic{table}}

\setcounter{section}{0}
\setcounter{subsection}{0}
\setcounter{subsubsection}{0}
\setcounter{paragraph}{0}
\setcounter{subparagraph}{0}
\setcounter{table}{0}
\setcounter{figure}{0}

\renewcommand{\thesection}{\Alph{section}}
\renewcommand{\thesubsection}{\thesection.\arabic{subsection}}
\renewcommand{\thesubsubsection}{\thesubsection.\arabic{subsubsection}}

\renewcommand{\theparagraph}{\thesubsubsection.\arabic{paragraph}}
\renewcommand{\thesubparagraph}{\theparagraph.\arabic{subparagraph}}

\makeatletter
\renewcommand{\@seccntformat}[1]{\csname the#1\endcsname.\quad}
\makeatother

\newcommand{\appsubsubsection}[1]{%
  \refstepcounter{subsubsection}%
  \par\medskip\noindent
  \textbf{\thesubsubsection\quad #1}\quad
}

\section{Appendix}

This Appendix contains additional details and data which were excluded from the conference version of this paper due to page limits.

\subsection{Experimental Setup and Preliminaries}
\appsubsubsection{Benchmark Instances}
\label{supp:sec:instance-selection}

This section provides the detailed construction procedure for the benchmark sets summarized in the main text.
Table~\ref{supp:tab:benchmark-summary} gives an overview of the candidate-pool and final-set sizes.

We used STRIPS planning tasks from the Autoscale benchmark suite \cite{Torralba_Seipp_Sievers_2021}, which are derived from International Planning Competition (IPC) benchmarks.

\paragraph{\texttt{blind}.}
We first constructed an initial candidate pool from the Autoscale benchmark suite by selecting at most one instance per domain such that RAM-based Fast Downward \astar\ solved the instance within 5 minutes and 8\,GiB, with search time of at least 10 seconds, at least $10^7$ expanded states, and at least 1\,GiB peak RAM usage.
This yielded 34 candidate instances.
From this pool, we selected 8 representative instances for the main \texttt{blind} experiments, prioritizing instances on which \siddastar exhibited relatively large memory usage and removing redundant cases with highly similar runtime and memory profiles.
The final \texttt{blind} set consists of \pddl{blocksworld-p23}, \pddl{data-network-p08}, \pddl{depots-p23}, \pddl{floortile-p03}, \pddl{mprime-p05}, \pddl{rovers-p03}, \pddl{snake-p22}, and \pddl{storage-p05}.

\paragraph{\texttt{blind-easy}.}
For the system-overhead experiments, we constructed a separate \texttt{blind-easy} set.
Candidate instances were required to be solvable by RAM-based \astar\ within 5 minutes and 8\,GiB, with search time of at most 1 second and at least $5\times10^5$ expanded states, again selecting at most one instance per domain.
This yielded 5 candidate instances, all of which were used:
\pddl{barman-p02}, \pddl{blocksworld-p04}, \pddl{gripper-p06}, \pddl{pegsol-p23}, and \pddl{termes-p01}.

\paragraph{\texttt{merge-and-shrink}.}
For the merge-and-shrink experiments, we first identified domains for which RAM-based \astar\ with merge-and-shrink solved at least one instance within 10 minutes and 8\,GiB, with at least $10^7$ expanded states and at least 1\,GiB peak RAM usage.
For each such domain, we retained the instance with the largest number of expanded states, yielding an initial candidate pool of 20 instances.
From this pool, we selected 8 representative instances, again prioritizing cases with relatively high \siddastar memory usage and removing redundant instances with highly similar runtime and memory profiles.
The final \texttt{merge-and-shrink} set consists of \pddl{agricola-p09}, \pddl{blocksworld-p10}, \pddl{data-network-p18}, \pddl{depots-p05}, \pddl{driverlog-p26}, \pddl{floortile-p07}, \pddl{hiking-p17}, and \pddl{zenotravel-p08}.

\begin{table}[t]
\centering
\small
\begin{tabular}{lcc}
\hline
Set & Candidate pool & Final set \\
\hline
\texttt{blind} & 34 & 8 \\
\texttt{blind-easy} & 5 & 5 \\
\texttt{merge-and-shrink} & 20 & 8 \\
\hline
\end{tabular}
\caption{Summary of benchmark-set construction: number of candidate instances after filtering and number of instances retained in the final benchmark set.}
\label{supp:tab:benchmark-summary}
\end{table}

\paragraph{Subsets for memory-pressure experiments in Section 5.1}
For the memory-pressure trend plots, we used small subsets drawn from the corresponding main benchmark sets.
These subsets were chosen so that both RAM-based \astar\ and \siddastar already consume several GiB of RAM in unrestricted runs, making the effects of reduced RAM and page-cache pressure easier to observe.
For \texttt{blind}, we used \pddl{blocksworld-p23}, \pddl{data-network-p08}, and \pddl{floortile-p03}.
For \texttt{merge-and-shrink}, we used \pddl{agricola-p09}, \pddl{data-network-p18}, and \pddl{floortile-p07}.

\paragraph{Hardware used for benchmark instance candidate filtering.}
Candidate selection for the \texttt{blind} and \texttt{blind-easy} sets was performed on the same machine as the main experiments, i.e., the Intel Core i7-14700KF platform with 32\,GiB RAM described in the main text.
Candidate filtering for the \texttt{merge-and-shrink} set was performed on a different machine equipped with an Intel Xeon E5-2670 v3 CPU. and 64\,GiB of %
RAM.
This auxiliary filtering step was used only to identify candidate instances under the stated per-run time and memory limits; all experimental results reported in the paper were obtained on the main experimental platform.

\subsection{Simple Immediate Duplicate Detection \astar}
\appsubsubsection{Additional Details of the open addressing Baseline}
\label{supp:sec:oa-details}

This section gives additional implementation details for \siddastar/OA, the open addressing baseline used in the main text.

\paragraph{Table structure.}
\siddastar/OA implements \closed as a fixed-size external hash table with $M=2^{27}$ slots.
The table is a power-of-two slot array and uses linear probing.
There is no deletion, no tombstones (because there is no deletion), and no resizing during search (because in our experiments, the initial table size was sufficiently large that load factor was low -- see below).
If all slots in the probe sequence are occupied, insertion fails; this did not occur in the reported experiments.

\paragraph{Record layout.}
Each slot stores exactly one complete \closed record consisting of:
(i) an index part containing an occupancy flag and a 32-bit hash value,
(ii) a fixed-size node header containing the parent pointer, generating operator, and packed $g$- and $h$-values, and
(iii) the packed state representation.
Thus, each slot stores both the duplicate-detection metadata and the full state payload.

\paragraph{Lookup and update behavior.}
For duplicate detection, the implementation first reads the index part of each probed slot.
If the slot is empty, lookup terminates.
If the stored hash matches the query hash, the full record is read and the packed state is compared exactly.
When a better duplicate is found, only the node header at the matching slot is updated.
When no match is found, insertion writes the complete record into the first empty slot in the probe sequence.

\paragraph{I/O behavior.}
The implementation uses no RAM-resident auxiliary index, no user-level caching, and no batched writes.
In the \texttt{pread}/\texttt{pwrite} version, probing first reads only the index part of each slot; the full record is read only on a hash match, and improved duplicates update only the node header in place.
Insertion writes the full record to the selected slot.
In the \texttt{mmap} version, the same logical access pattern is implemented through memory-mapped pages.
Thus, newly inserted \closed records incur random full-record writes to probe-determined locations.

\paragraph{Load factor.}
To clarify whether the poor performance of \siddastar/OA is simply due to a high table load, Table~\ref{supp:tab:oa-load-factor} reports the load factor at the 600-second cutoff for the RAM-limited runs used in the main comparison.
The observed values are low across the benchmark set: even the largest value is below $0.05$.
Thus, the poor performance of \siddastar/OA is not explained by near-full-table behavior alone.

\begin{table}[htb]
\centering
\small
\setlength{\tabcolsep}{4pt}
\renewcommand{\arraystretch}{1.1}
\begin{tabular}{l|rr}
\hline
Problem & \siddastarpwrite/OA (\%) & \siddastarmmap/OA (\%) \\
\hline
\multicolumn{3}{l}{{\bf Blind heuristic}}\\
\hline
\pddl{blocksworld-p23}  & 4.63 & 1.31 \\
\pddl{data-network-p08} & 1.49 & 0.36 \\
\pddl{depots-p23}       & 1.85 & 0.51 \\
\pddl{floortile-p03}    & 2.50 & 0.67 \\
\pddl{mprime-p05}       & 2.53 & 0.80 \\
\pddl{rovers-p03}       & 2.65 & 0.81 \\
\pddl{snake-p22}        & 3.35 & 1.18 \\
\pddl{storage-p05}      & 3.85 & 1.50 \\
\hline
\multicolumn{3}{l}{{\bf Merge-and-shrink heuristic}}\\
\hline
\pddl{agricola-p09}     & 4.82 & 1.75 \\
\pddl{blocksworld-p10}  & 3.47 & 1.50 \\
\pddl{data-network-p18} & 1.44 & 0.53 \\
\pddl{depots-p05}       & 1.96 & 0.74 \\
\pddl{driverlog-p26}    & 2.89 & 1.04 \\
\pddl{floortile-p07}    & 2.83 & 0.78 \\
\pddl{hiking-p17}       & 2.01 & 0.90 \\
\pddl{zenotravel-p08}   & 2.22 & 0.69 \\
\hline
\end{tabular}
\caption{Load factor of \siddastar/OA at the 600-second cutoff for the RAM-limited runs used in the main comparison. Values are percentages of occupied slots out of the total number of slots ($2^{27}$).}
\label{supp:tab:oa-load-factor}
\end{table}

\subsection{Experimental Evaluation}
\appsubsubsection{Additional Details}
\label{supp:sec:experiments-additional-details}

Parameters for the merge-and-shrink heuristic were:
\begin{quote}
\footnotesize
\noindent\parbox{\linewidth}{%
\raggedright
merge\_and\_shrink(\\
\hspace*{1em}shrink\_strategy=shrink\_bisimulation(greedy=false),\\
\hspace*{1em}merge\_strategy=merge\_sccs(\\
\hspace*{2em}order\_of\_sccs=topological,\\
\hspace*{2em}merge\_selector=score\_based\_filtering(\\
\hspace*{3em}scoring\_functions=[goal\_relevance,\\
\hspace*{4em}dfp, total\_order])),\\
\hspace*{1em}label\_reduction=exact(\\
\hspace*{2em}before\_shrinking=true, before\_merging=false),\\
\hspace*{1em}max\_states=50000,\\
\hspace*{1em}threshold\_before\_merge=1)%
}
\end{quote}

\appsubsubsection{Preallocation of files}
\label{supp:sec:preallocation}

To reduce variability caused by file growth during search, \siddastar preallocates, using \texttt{fallocate}, the single external file used to store the closed list at the beginning of the first run.
In contrast, the open list consists of multiple bucket files, and these files are not preallocated.

Preallocation reduces file-system metadata updates and mitigates performance noise due to incremental extent allocation and possible fragmentation as the file grows.
In our environment, the preallocation overhead is negligible: even for a file on the order of 100\,GiB, \texttt{fallocate} completes in well under one second.
Since this one-time setup cost is amortized over all runs and  not part of the  search procedure, we exclude the corresponding wall-clock time from the reported search time.

Note that preallocation only allocates space and does \textit{not} write any data. \siddastar does not require the external file to be initialized (e.g., filled with zeroes), so a single large external file only needs to be preallocated once and can be reused for many runs on different problems.

\appsubsubsection{Nodes Expanded by Eager vs. Lazy Duplicate Detection}
\label{supp:sec:eager-vs-lazy-dd}

\ifdefined\verifiedexpansions  %

All of the evaluated implementations (\astar, \siddastarpwrite, \siddastarmmap, A*-IDD) are derived from the Fast Downward planner code, and use the same successor generation function, same tie-breaking policy (by lexicographic order of ($f$,$h$,FIFO), where FIFO is  First-In-First-Out order of entry into the ($f$,$h$) bucket of \open), and the same heuristic function.
The only notable difference  is when duplicate detection is performed:
the standard Fast Downward \astar performs eager duplicate detection at node generation time, whereas all other variants perform lazy duplicate detection and defer duplicate detection until node extraction.

We executed all variants on all benchmark instances without RAM limitation (for maximum speed). All algorithms solved all instances. 
We verified that they all expanded the same number of nodes on all instances, shown in Table \ref{tab:eager-vs-lazy-expanded}.
Thus, expansion rates are an appropriate metric of comparison, as we are concerned with how rapidly each method processes the (same) work.

\else 

To empirically verify that lazy duplicate detection does not change the expansion behavior of \astar on the benchmark instances used in this paper, we implemented a lazy-DD variant of the standard Fast Downward \astar planner.

Both the standard Fast Downward \astar implementation and this lazy-DD variant use the same tie-breaking rule: nodes are ordered by increasing $f$, then by increasing $h$, and then by FIFO order according to when the node enters the $\open$ bucket.
The only difference between them is when duplicate detection is performed:
the standard Fast Downward \astar performs eager duplicate detection at node generation time, whereas the verification variant defers duplicate detection until node extraction.

Table~\ref{tab:eager-vs-lazy-expanded} reports the number of expanded nodes for the benchmark instances used in our experiments.
For all instances in these benchmark sets, the eager-DD and lazy-DD variants expanded exactly the same number of nodes.
\fi

\ifdefined\verifiedexpansions

\begin{table}[htb] %
\centering
\small
\setlength{\tabcolsep}{6pt}
\renewcommand{\arraystretch}{1.1}

\begin{tabular}{l|r}
\hline
Problem & Nodes Expanded \\\hline
\multicolumn{2}{l}{{\bf Blind}}\\
\hline
\pddl{blocksworld-p23} & 70{,}971{,}548\\
\pddl{data-network-p08} & 22{,}736{,}108\\
\pddl{depots-p23} & 52{,}179{,}666 \\
\pddl{floortile-p03} & 54{,}901{,}766 \\
\pddl{mprime-p05} & 24{,}565{,}932 \\
\pddl{rovers-p03} & 101{,}909{,}693 \\
\pddl{snake-p22} & 76{,}835{,}143 \\
\pddl{storage-p05} & 50{,}574{,}701 \\
\hline
\multicolumn{2}{l}{{\bf Blind-easy}}\\
\hline
\pddl{barman-p02} & 514{,}830 \\
\pddl{blocksworld-p04} & 616{,}846 \\
\pddl{gripper-p06} & 868{,}326 \\
\pddl{pegsol-p23} & 563{,}522 \\
\pddl{termes-p01} & 522{,}686 \\
\hline
\multicolumn{2}{l}{{\bf Merge-and-shrink}}\\
\hline
\pddl{agricola-p09} & 50{,}221{,}643 \\
\pddl{blocksworld-p10} & 18{,}660{,}540 \\
\pddl{data-network-p18} & 23{,}085{,}944 \\
\pddl{depots-p05} & 37{,}558{,}946 \\
\pddl{driverlog-p26} & 15{,}322{,}079 \\
\pddl{floortile-p07} & 68{,}218{,}082 \\
\pddl{hiking-p17} & 92{,}609{,}121 \\
\pddl{zenotravel-p08} & 12{,}260{,}839 \\
\hline
\end{tabular}
\caption{Number of expanded nodes to solve benchmark instances. All algorithms expanded the same number of nodes.}
\label{tab:eager-vs-lazy-expanded} %
\end{table}

\else
\begin{table}[htb] 
\centering
\small
\setlength{\tabcolsep}{6pt}
\renewcommand{\arraystretch}{1.1}

\begin{tabular}{l|rr}
\hline
Problem & \astar (eager DD) & \astar (lazy DD) \\
\hline
\multicolumn{3}{l}{{\bf Blind}}\\
\hline
\pddl{blocksworld-p23} & 70{,}971{,}548 & 70{,}971{,}548 \\
\pddl{data-network-p08} & 22{,}736{,}108 & 22{,}736{,}108 \\
\pddl{depots-p23} & 52{,}179{,}666 & 52{,}179{,}666 \\
\pddl{floortile-p03} & 54{,}901{,}766 & 54{,}901{,}766 \\
\pddl{mprime-p05} & 24{,}565{,}932 & 24{,}565{,}932 \\
\pddl{rovers-p03} & 101{,}909{,}693 & 101{,}909{,}693 \\
\pddl{snake-p22} & 76{,}835{,}143 & 76{,}835{,}143 \\
\pddl{storage-p05} & 50{,}574{,}701 & 50{,}574{,}701 \\
\hline
\multicolumn{3}{l}{{\bf Blind-easy}}\\
\hline
\pddl{barman-p02} & 514{,}830 & 514{,}830 \\
\pddl{blocksworld-p04} & 616{,}846 & 616{,}846 \\
\pddl{gripper-p06} & 868{,}326 & 868{,}326 \\
\pddl{pegsol-p23} & 563{,}522 & 563{,}522 \\
\pddl{termes-p01} & 522{,}686 & 522{,}686 \\
\hline
\multicolumn{3}{l}{{\bf Merge-and-shrink}}\\
\hline
\pddl{agricola-p09} & 50{,}221{,}643 & 50{,}221{,}643 \\
\pddl{blocksworld-p10} & 18{,}660{,}540 & 18{,}660{,}540 \\
\pddl{data-network-p18} & 23{,}085{,}944 & 23{,}085{,}944 \\
\pddl{depots-p05} & 37{,}558{,}946 & 37{,}558{,}946 \\
\pddl{driverlog-p26} & 15{,}322{,}079 & 15{,}322{,}079 \\
\pddl{floortile-p07} & 68{,}218{,}082 & 68{,}218{,}082 \\
\pddl{hiking-p17} & 92{,}609{,}121 & 92{,}609{,}121 \\
\pddl{zenotravel-p08} & 12{,}260{,}839 & 12{,}260{,}839 \\
\hline
\end{tabular}
\caption{Number of expanded nodes for standard \astar with eager duplicate detection and \astar with lazy duplicate detection.}
\label{tab:eager-vs-lazy-expanded}
\end{table}
\fi

\appsubsubsection{A*-IDD Comparison of FIFO vs LIFO Tie-Breaking}
\label{supp:sec:a*-idd-fifo-vs-lifo-tiebreak}

Table~\ref{supp:tab:astar-idd-fifo-lifo-full-expansion-rate} compares expansion rates for  A*-IDD with FIFO and LIFO tie-breaking under a 1\,GiB RAM limit.
Each run was executed with a 600-second time limit, and the table reports the average expansion rate over the full run.

The purpose of this comparison is not to claim that the two variants are search-equivalent.
Because tie-breaking affects the order in which nodes are expanded within an $f$-layer, FIFO and LIFO may expand different numbers of nodes and may solve different subsets of instances within the time limit.

Nevertheless, the results show that their overall expansion rates are broadly similar across our benchmark instances.
In some cases FIFO is faster, while in others LIFO is faster, and neither dominates. We do not observe a consistent or substantial performance gap favoring one policy over the other.
Therefore, in the experiments in the paper, we use the FIFO policy, so that we can apply a uniform  tie-breaking policy among all of the algorithms.

\begin{table}[htb]
\centering
\small
\setlength{\tabcolsep}{6pt}
\renewcommand{\arraystretch}{1.1}

\begin{tabular}{l|rr}
\hline
Problem & A*-IDD (FIFO) & A*-IDD (LIFO) \\
\hline
\multicolumn{3}{l}{{\bf Blind heuristic}}\\
\hline
\pddl{blocksworld-p23} & 74{,}784 & 86{,}769$^{*}$ \\
\pddl{data-network-p08} & 16{,}005 & 21{,}438 \\
\pddl{depots-p23} & 22{,}461 & 18{,}760 \\
\pddl{floortile-p03} & 38{,}972 & 31{,}973 \\
\pddl{mprime-p05} & 26{,}715 & 21{,}080 \\
\pddl{rovers-p03} & 32{,}912 & 28{,}866 \\
\pddl{snake-p22} & 54{,}767 & 55{,}671 \\
\pddl{storage-p05} & 56{,}666 & 42{,}553 \\

\hline
\multicolumn{3}{l}{{\bf Merge-and-shrink heuristic}}\\
\hline
\pddl{agricola-p09}     & 26{,}805 & 24{,}631 \\
\pddl{blocksworld-p10}  & 52{,}350$^{*}$ & 48{,}223$^{*}$ \\
\pddl{data-network-p18} & 11{,}306 & 11{,}006 \\
\pddl{depots-p05}       & 18{,}157 & 17{,}620 \\
\pddl{driverlog-p26}    & 31{,}800$^{*}$ & 32{,}270$^{*}$ \\
\pddl{floortile-p07}    & 36{,}634 & 33{,}934 \\
\pddl{hiking-p17}       & 23{,}392 & 24{,}123 \\
\pddl{zenotravel-p08}   & 20{,}003 & 21{,}039$^{*}$ \\
\hline
\end{tabular}

\caption{Average expansion rates (expansions/sec) over the full 600-second run under the 1\,GiB RAM limit for A*-IDD with FIFO and LIFO tie-breaking. Entries marked with $^{*}$ indicate that a solution was found within the time limit.}
\label{supp:tab:astar-idd-fifo-lifo-full-expansion-rate}
\end{table}

\appsubsubsection{Additional Figures and Tables}

This subsection provides additional measurements corresponding to the experiments in the main text.

Table~\ref{supp:tab:combined-full-expansion-rate} reports \emph{full expansion rates} for the same 600-second memory-pressure experiments summarized in Section~5.1 of the main text. %
That is, each entry is the average expansion rate over the entire run.
In contrast, the main text reports the expansion rate measured over the last 60 seconds of each 600-second run.

The reason for emphasizing the last-60-seconds expansion rate in the main text is that our goal there is to characterize algorithm performance under sustained high memory pressure, after phenomena such as swap thrashing and page-cache eviction have become significant.
A full-run average can be strongly influenced by the earlier phase of the run, before such effects fully manifest, and therefore can obscure the behavior of most interest in the memory-pressure setting.
For this reason, we report the last-60-seconds metric in the main text and provide the corresponding full-run averages here as supplementary information.

Table~\ref{tab:expansion-rate-ram-by-method-mands} complements Section~5.2 of the main text by reporting the no-memory-limit results for the \texttt{merge-and-shrink} benchmark set, analogous to the \texttt{blind}-heuristic table shown in the main text.

Figures~\ref{supp:fig:blind-ram-limit} and \ref{supp:fig:merge-and-shrink-ram-limit} show running average expansion-rate curves for the memory-pressure experiments, Figures~\ref{supp:fig:raw-blind-ram-limit} and \ref{supp:fig:raw-merge-and-shrink-ram-limit} show the corresponding raw expansion-rate curves, and Figures~\ref{supp:fig:total-blind-ram-limit} and \ref{supp:fig:total-merge-and-shrink-ram-limit} show total expansions as a function of cumulative wall-clock time.
All curves are based on nominal 5-second sampling; during stall periods, sampling may be delayed, so some plotted values reflect expansions accumulated over intervals longer than 5 seconds.

Some values in Table~\ref{supp:tab:combined-full-expansion-rate} may differ from the running average expansion-rate values that can be visually inferred from the figures, even when they refer to the same instance and nominal RAM limit.
This is because the table and figure data were obtained from separate runs of the same experimental configuration, rather than from exactly the same execution logs.
Under memory pressure, run-to-run variation is non-negligible, especially because performance is sensitive to page-cache behavior and related OS-level effects.
Accordingly, small discrepancies between the tabulated values and plotted curves should be expected.

\begin{table*}[htb]
\centering
\small
\setlength{\tabcolsep}{4pt}
\renewcommand{\arraystretch}{1.1}

\begin{tabular}{l|rrGrrGrr|rr}
\hline
Problem & \multicolumn{6}{c|}{{\bf 1.0GiB RAM limit}} & \multicolumn{2}{c}{{\bf 600MiB RAM limit}} \\

& \astar
& A*-IDD
& \siddastarpwrite
& \siddastarmmap
& \siddastarpwrite
& \siddastarmmap
& A*-IDD
& \siddastarpwrite \\
&
&
& Sep. Chaining
& Sep. Chaining
& Open Address.
& Open Address.
&
& Sep. Chaining \\
\hline
\multicolumn{9}{l}{{\bf Blind heuristic}}\\
\hline
\pddl{blocksworld-p23} & 66{,}078 & \textbf{74{,}784} & 73{,}378 & 59{,}005 & 10{,}357 & 2{,}939 & 49{,}527 & \textbf{53{,}569} \\
\pddl{data-network-p08} & 22{,}324 & 16{,}005 & 44{,}929$^{*}$ & \textbf{54{,}696}$^{*}$ & 3{,}353 & 809 & 8{,}735 & \textbf{24{,}724} \\
\pddl{depots-p23} & \textbf{72{,}695} & 22{,}461 & 58{,}579 & 49{,}840 & 4{,}132 & 1{,}131 & 9{,}307 & \textbf{28{,}112} \\
\pddl{floortile-p03} & 39{,}848 & 38{,}972 & \textbf{64{,}113} & 56{,}711 & 5{,}581 & 1{,}509 & 18{,}303 & \textbf{35{,}316} \\
\pddl{mprime-p05} & 51{,}711$^{*}$ & 26{,}715 & 70{,}357$^{*}$ & \textbf{76{,}983}$^{*}$ & 5{,}663 & 1{,}783 & 13{,}970 & \textbf{32{,}215} \\
\pddl{rovers-p03} & \textbf{98{,}154} & 32{,}912 & 65{,}579 & 58{,}191 & 5{,}935 & 1{,}814 & 21{,}136 & \textbf{43{,}686} \\
\pddl{snake-p22} & \textbf{79{,}876} & 54{,}767 & 69{,}207 & 53{,}541 & 7{,}484 & 2{,}644 & 31{,}191 & \textbf{51{,}055} \\
\pddl{storage-p05} & 67{,}310 & 56{,}666 & \textbf{71{,}340} & 56{,}502 & 8{,}621 & 3{,}352 & 30{,}364 & \textbf{45{,}718} \\

\hline
\multicolumn{9}{l}{{\bf Merge-and-shrink heuristic}}\\
\hline
\pddl{agricola-p09}     & \textbf{116{,}525}$^{*}$ & 26{,}805 & 63{,}982 & 56{,}146 & 10{,}768 & 3{,}916 & 22{,}541 & \textbf{48{,}305} \\
\pddl{blocksworld-p10}  & 101{,}822$^{*}$ & 52{,}350$^{*}$ & \textbf{167{,}735}$^{*}$ & 131{,}153$^{*}$ & 7{,}762 & 3{,}357 & 40{,}048$^{*}$ & \textbf{97{,}076}$^{*}$ \\
\pddl{data-network-p18} & 23{,}255 & 11{,}306 & 33{,}439 & \textbf{36{,}303} & 3{,}207 & 1{,}195 & 7{,}547 & \textbf{21{,}202} \\
\pddl{depots-p05}       & 44{,}503 & 18{,}157 & \textbf{49{,}003} & 44{,}734 & 4{,}382 & 1{,}657 & 13{,}800 & \textbf{31{,}519} \\
\pddl{driverlog-p26}    & 15{,}537 & 31{,}800$^{*}$ & \textbf{104{,}873}$^{*}$ & 101{,}913$^{*}$ & 6{,}472 & 3{,}321 & 21{,}602 & \textbf{53{,}187}$^{*}$ \\
\pddl{floortile-p07}    & 32{,}758 & 36{,}634 & \textbf{63{,}742} & 48{,}817 & 6{,}325 & 1{,}749 & 19{,}560 & \textbf{39{,}812} \\
\pddl{hiking-p17}       & \textbf{144{,}546} & 23{,}392 & 57{,}146 & 62{,}845 & 4{,}495 & 2{,}010 & 19{,}465 & \textbf{39{,}111} \\
\pddl{zenotravel-p08}   & 18{,}559 & 20{,}003 & 64{,}370$^{*}$ & \textbf{70{,}550}$^{*}$ & 4{,}957 & 1{,}546 & 15{,}797 & \textbf{36{,}491}$^{*}$ \\
\hline
\end{tabular}

\caption{Full expansion rates (expansions/sec) for the blind-heuristic and merge-and-shrink benchmark sets. Entries marked with $^{*}$ indicate that a solution was found.}
\label{supp:tab:combined-full-expansion-rate}
\end{table*}

\begin{table}[htb]
\centering
\scriptsize
\setlength{\tabcolsep}{3pt}
\renewcommand{\arraystretch}{1.1}
\resizebox{\columnwidth}{!}{%
\begin{tabular}{l|rr|rr|rr}
\hline
Problem
& \multicolumn{2}{c|}{\astar}
& \multicolumn{2}{c|}{\siddastarmmap}
& \multicolumn{2}{c}{\siddastarpwrite} \\
& Exp.\ rate & RAM
& Exp.\ rate & RAM
& Exp.\ rate & RAM \\
\hline
\pddl{agricola-p09}
  & \textbf{380{,}295} & \textbf{3{,}860}
  & 97{,}350  & 6{,}311
  & 84{,}017  & 6{,}294 \\
\pddl{blocksworld-p10}
  & \textbf{368{,}343} & \textbf{3{,}275}
  & 146{,}436 & 3{,}428
  & 181{,}922 & 3{,}409 \\
\pddl{data-network-p18}
  & \textbf{161{,}333} & \textbf{4{,}379}
  & 46{,}494  & 8{,}729
  & 36{,}453  & 8{,}647 \\
\pddl{depots-p05}
  & \textbf{283{,}293} & \textbf{3{,}775}
  & 66{,}160  & 12{,}110
  & 61{,}509  & 12{,}071 \\
\pddl{driverlog-p26}
  & \textbf{202{,}338} & 6{,}224
  & 109{,}347 & 5{,}080
  & 113{,}520 & \textbf{5{,}047} \\
\pddl{floortile-p07}
  & \textbf{386{,}153} & \textbf{6{,}174}
  & 102{,}173 & 6{,}471
  & 137{,}029 & 6{,}233 \\
\pddl{hiking-p17}
  & \textbf{609{,}958} & \textbf{3{,}211}
  & 176{,}817 & 5{,}449
  & 69{,}296  & 5{,}431 \\
\pddl{zenotravel-p08}
  & \textbf{187{,}454} & \textbf{3{,}412}
  & 73{,}053  & 6{,}351
  & 67{,}237  & 6{,}327 \\
\hline
\end{tabular}%
}
\caption{Expansion rates (states/sec) and peak RAM usage (cgroup, MiB) on 8 instances (merge-and-shrink heuristic) with no RAM limit (i.e., full usage of the 32\,GB system RAM).}
\label{tab:expansion-rate-ram-by-method-mands}
\end{table}

\begin{figure*}[htb]
  \centering

  \begin{subfigure}{0.32\textwidth}
    \centering
    \includegraphics[width=\linewidth]{figures/blocksworld_4GiB_blind.png}
    \par\vspace{0.4em}
    \includegraphics[width=\linewidth]{figures/blocksworld_2GiB_blind.png}
    \par\vspace{0.4em}
    \includegraphics[width=\linewidth]{figures/blocksworld_1GiB_blind.png}
    \caption{\pddl{blocksworld-p23}}
    \label{fig:blocksworld-blind-ram-limit}
  \end{subfigure}
  \hfill
  \begin{subfigure}{0.32\textwidth}
    \centering
    \includegraphics[width=\linewidth]{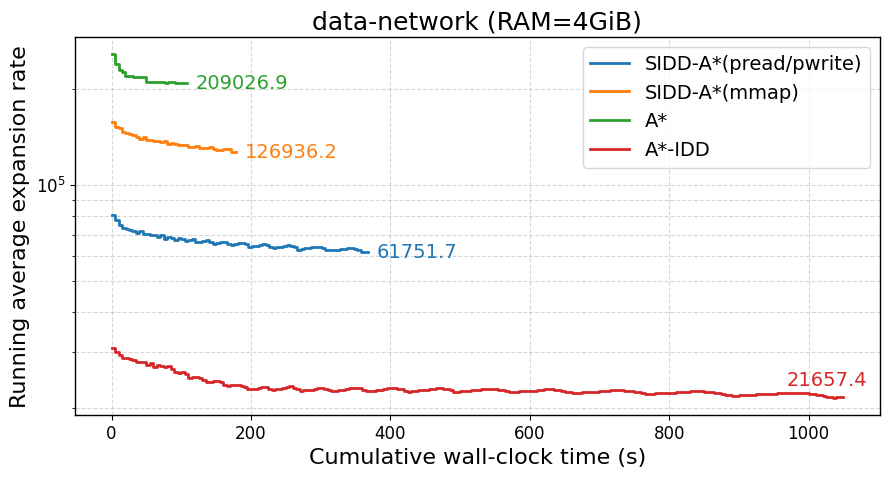}
    \par\vspace{0.4em}
    \includegraphics[width=\linewidth]{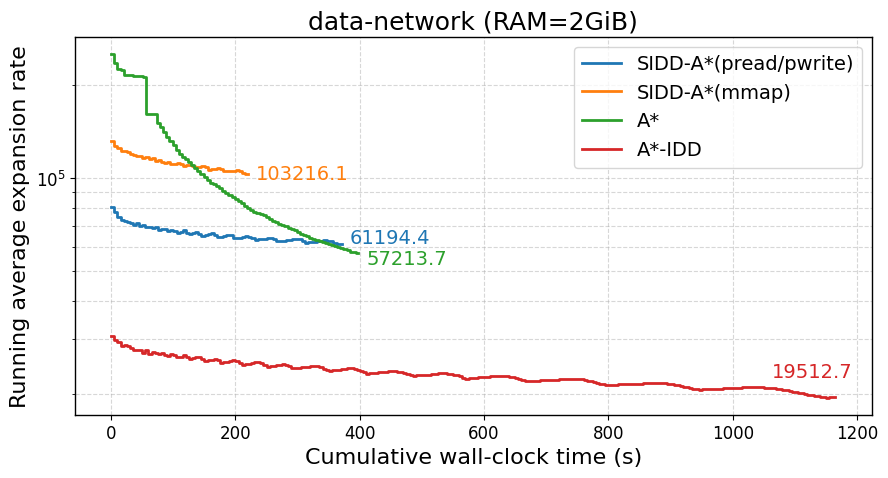}
    \par\vspace{0.4em}
    \includegraphics[width=\linewidth]{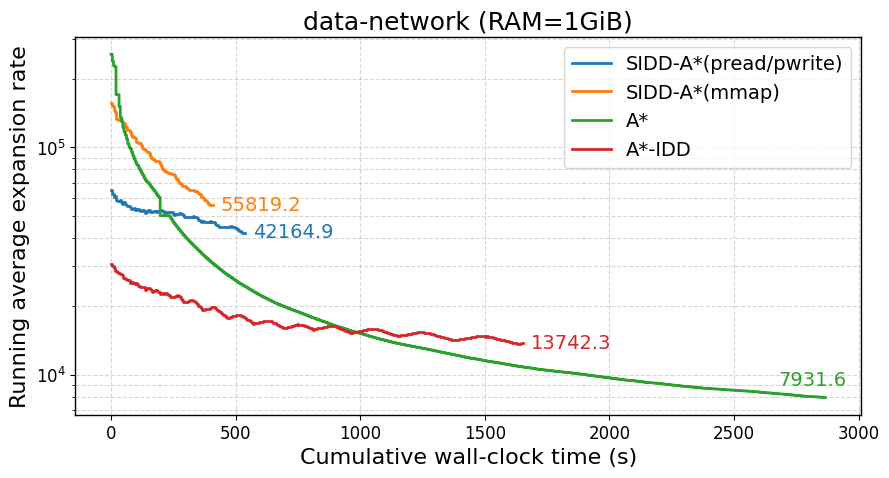}
    \caption{\pddl{data-network-p08}}
    \label{fig:data-network-blind-ram-limit}
  \end{subfigure}
  \hfill
  \begin{subfigure}{0.32\textwidth}
    \centering
    \includegraphics[width=\linewidth]{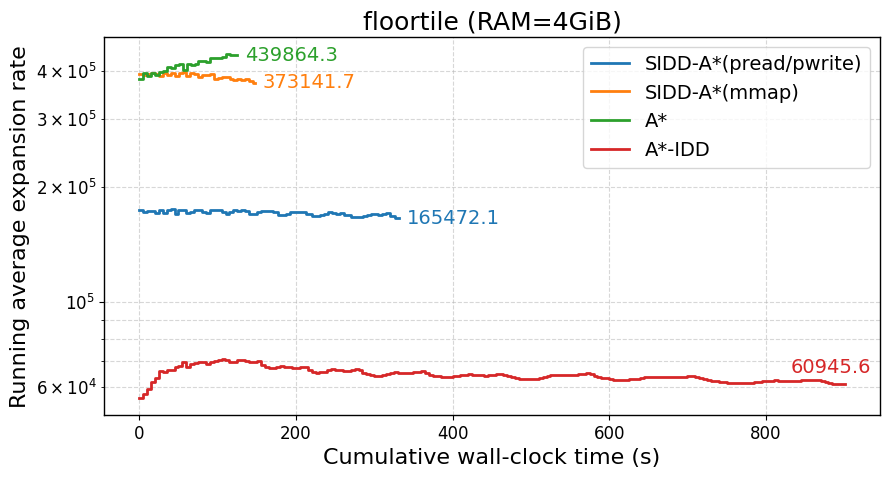}
    \par\vspace{0.4em}
    \includegraphics[width=\linewidth]{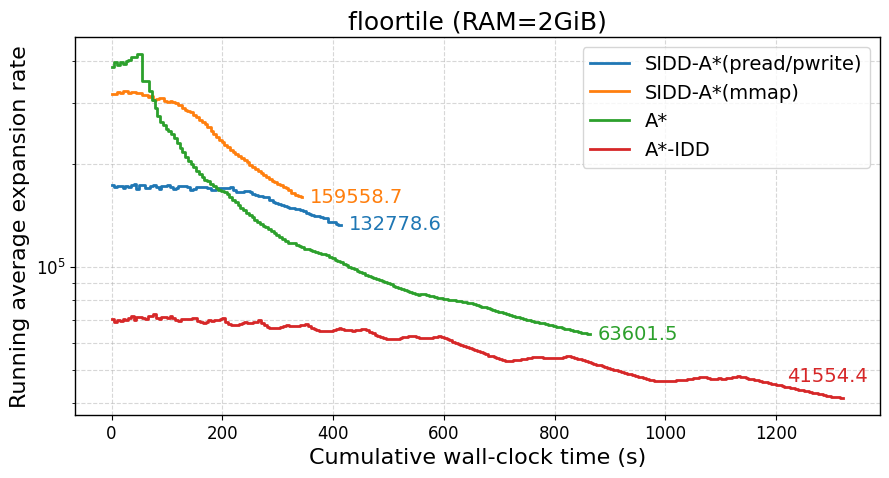}
    \par\vspace{0.4em}
    \includegraphics[width=\linewidth]{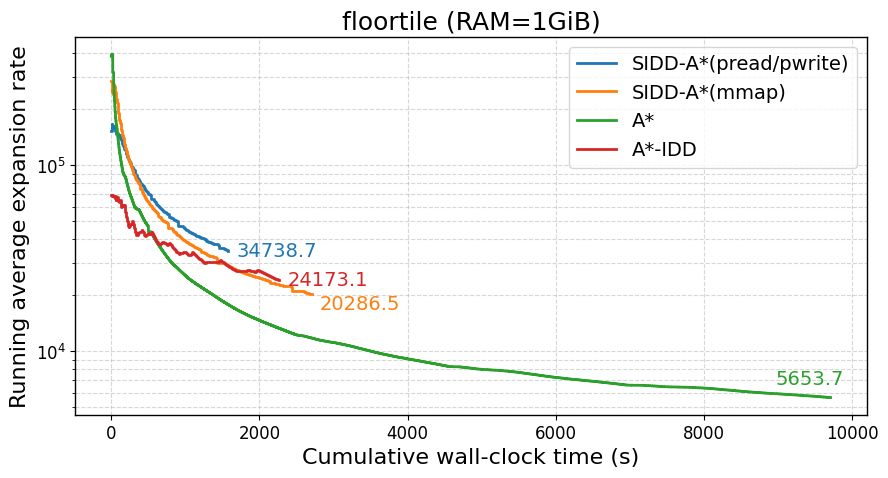}
    \caption{\pddl{floortile-p03}}
    \label{fig:floortile-blind-ram-limit}
  \end{subfigure}

  \caption{RAM limitation under the \texttt{blind} heuristic. Each plot shows running average expansion rate as a function of cumulative wall-clock time. Each column shows results for RAM limits of 4\,GiB, 2\,GiB, and 1\,GiB from top to bottom.}
  \label{supp:fig:blind-ram-limit}
\end{figure*}

\begin{figure*}[htb]
  \centering

  \begin{subfigure}{0.32\textwidth}
    \centering
    \includegraphics[width=\linewidth]{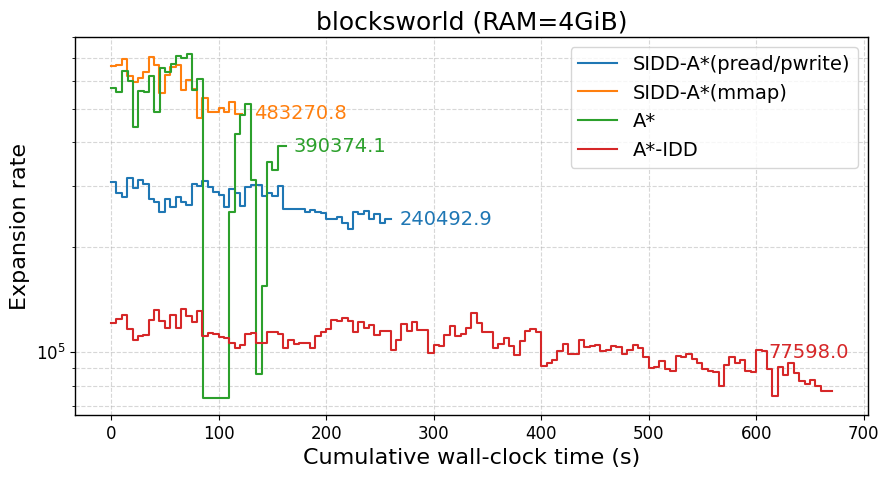}
    \par\vspace{0.4em}
    \includegraphics[width=\linewidth]{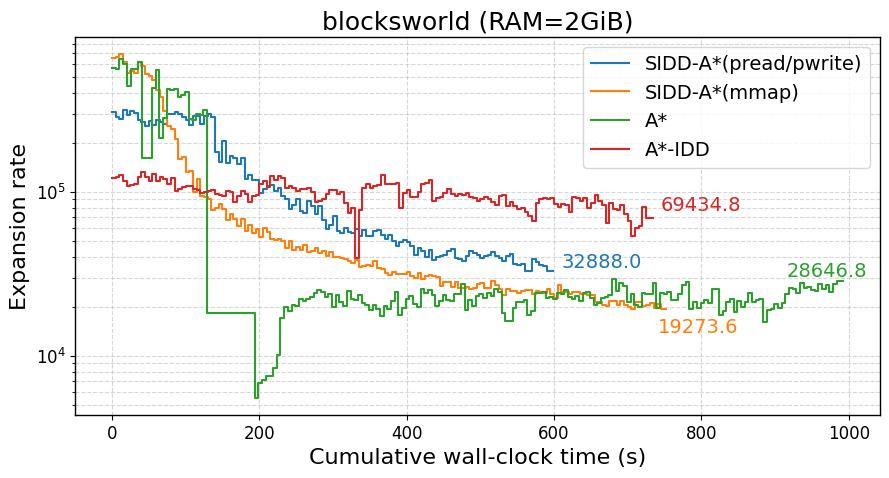}
    \par\vspace{0.4em}
    \includegraphics[width=\linewidth]{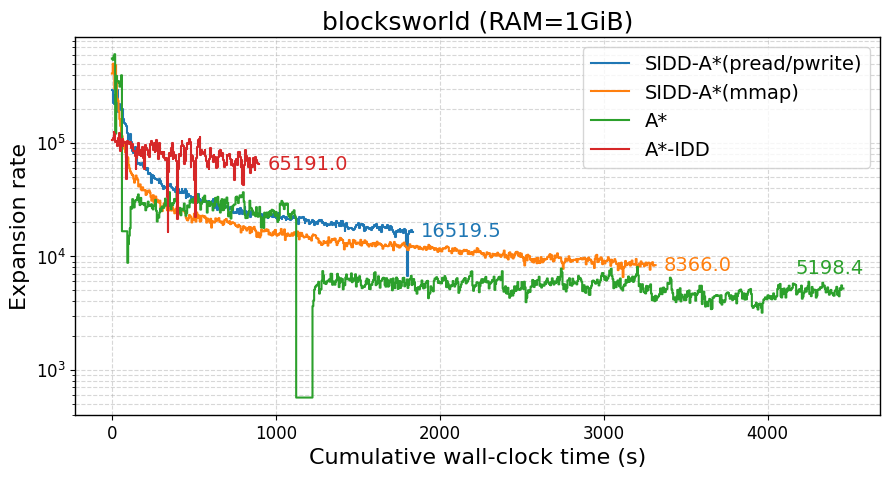}
    \caption{\pddl{blocksworld-p23}}
    \label{fig:raw-blocksworld-blind-ram-limit}
  \end{subfigure}
  \hfill
  \begin{subfigure}{0.32\textwidth}
    \centering
    \includegraphics[width=\linewidth]{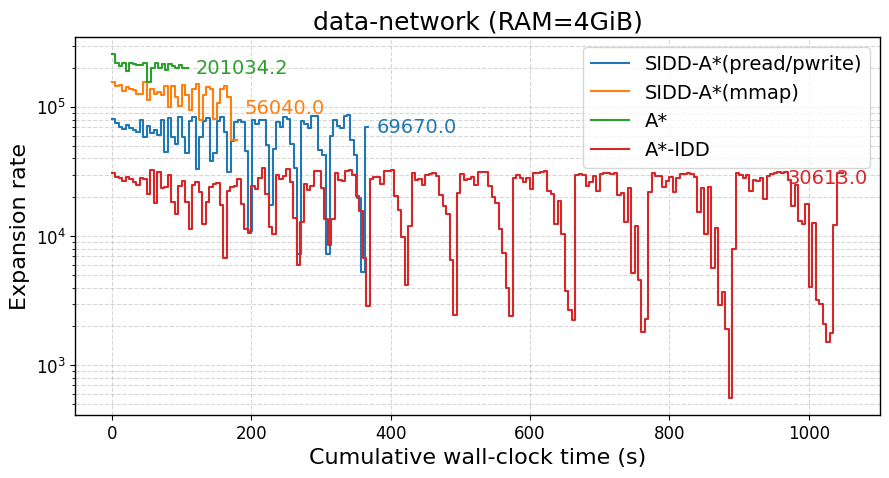}
    \par\vspace{0.4em}
    \includegraphics[width=\linewidth]{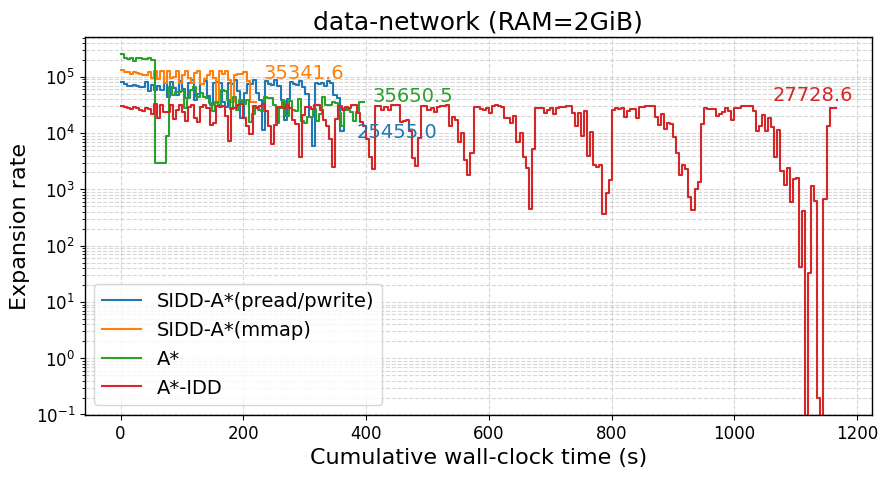}
    \par\vspace{0.4em}
    \includegraphics[width=\linewidth]{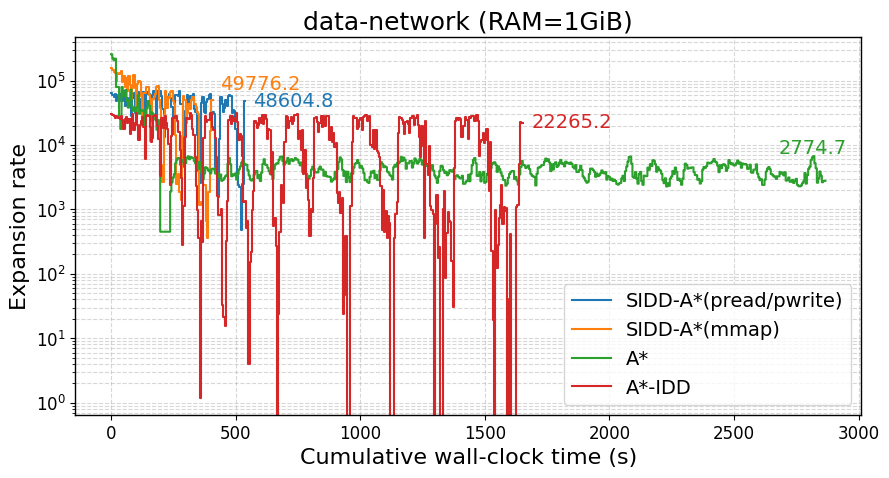}
    \caption{\pddl{data-network-p08}}
    \label{fig:raw-data-network-blind-ram-limit}
  \end{subfigure}
  \hfill
  \begin{subfigure}{0.32\textwidth}
    \centering
    \includegraphics[width=\linewidth]{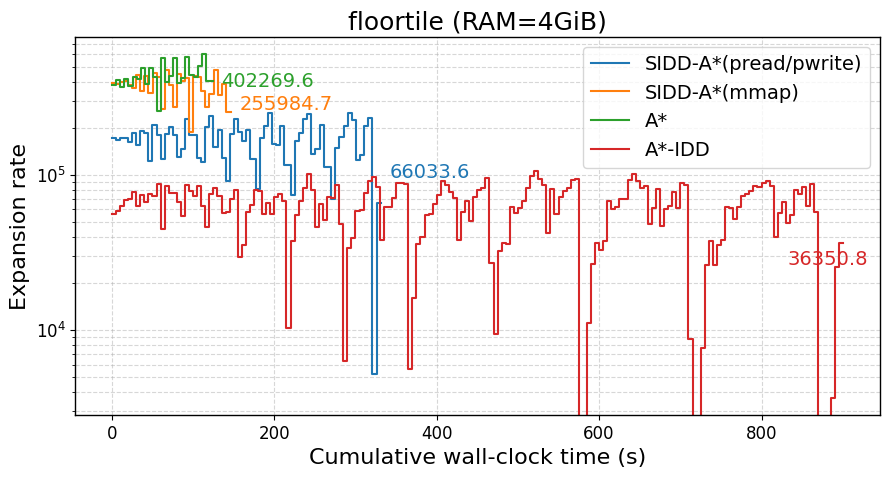}
    \par\vspace{0.4em}
    \includegraphics[width=\linewidth]{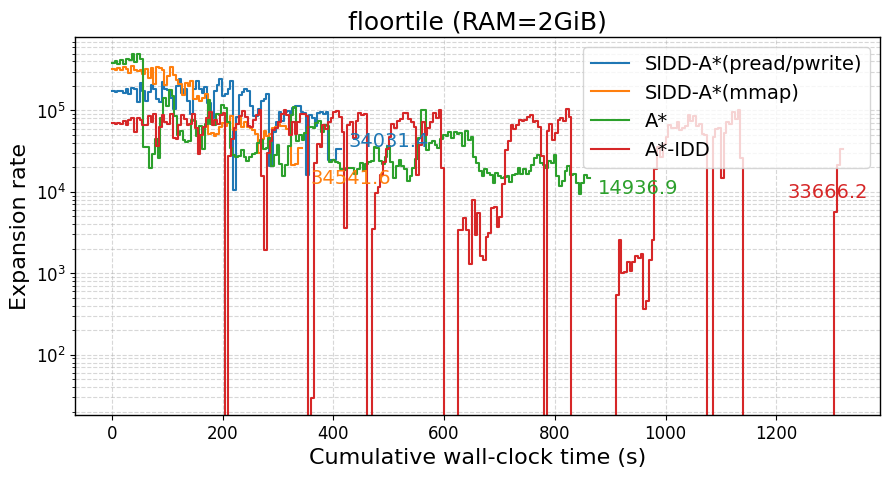}
    \par\vspace{0.4em}
    \includegraphics[width=\linewidth]{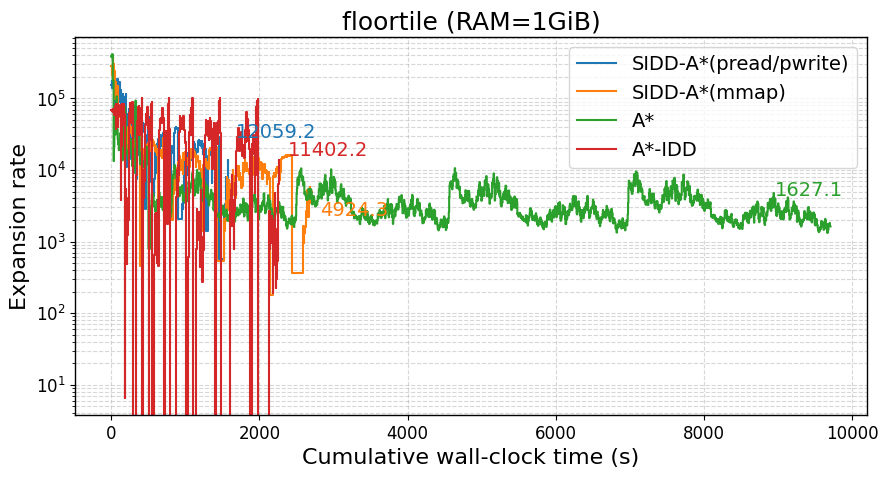}
    \caption{\pddl{floortile-p03}}
    \label{fig:raw-floortile-blind-ram-limit}
  \end{subfigure}

  \caption{RAM limitation under the \texttt{blind} heuristic. Each plot shows expansion rate as a function of cumulative wall-clock time. Each column shows results for RAM limits of 4\,GiB, 2\,GiB, and 1\,GiB from top to bottom.}
  \label{supp:fig:raw-blind-ram-limit}
\end{figure*}

\begin{figure*}[htb]
  \centering

  \begin{subfigure}{0.31\textwidth}
    \centering
    \includegraphics[width=\linewidth]{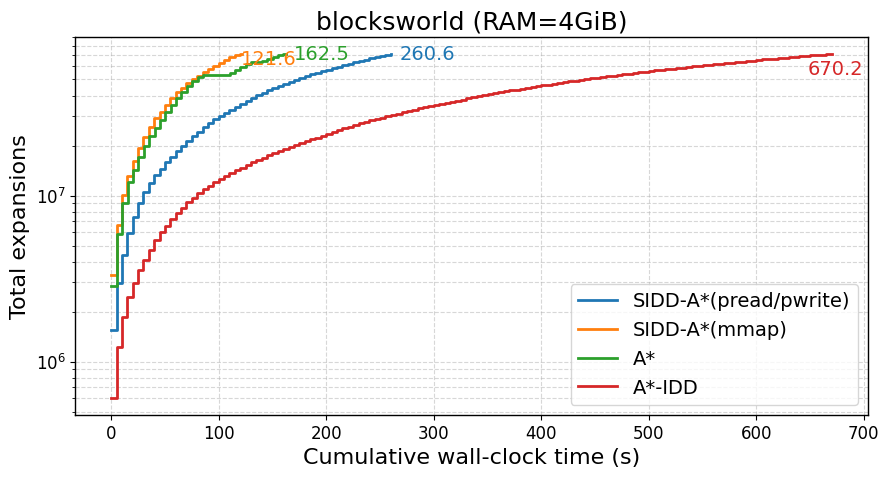}
    \par\vspace{0.4em}
    \includegraphics[width=\linewidth]{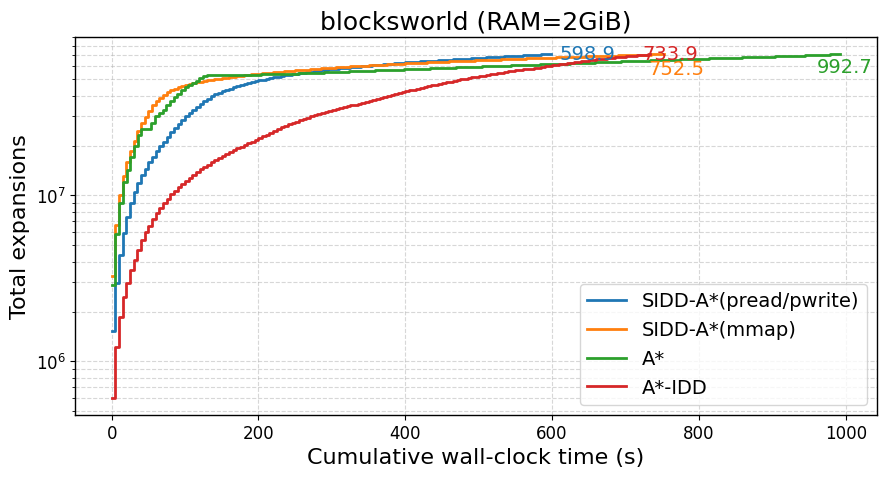}
    \par\vspace{0.4em}
    \includegraphics[width=\linewidth]{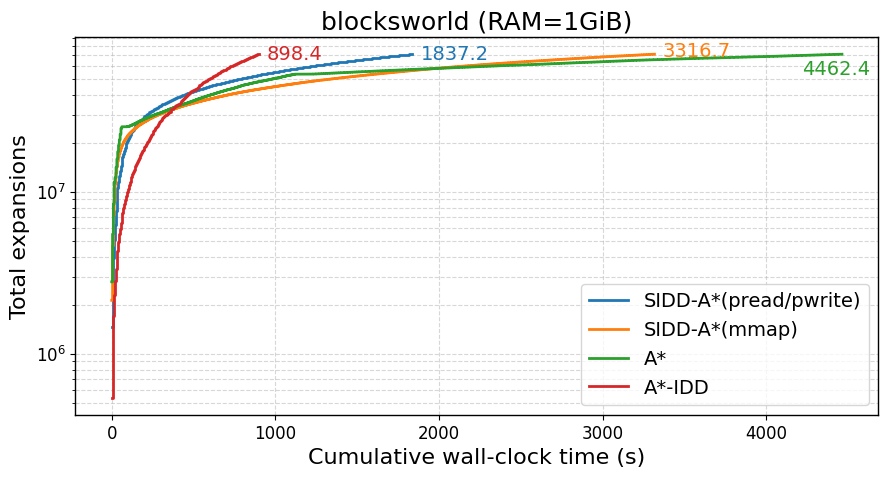}
    \caption{\pddl{blocksworld-p23}}
    \label{fig:total-blocksworld-blind-ram-limit}
  \end{subfigure}
  \hfill
  \begin{subfigure}{0.31\textwidth}
    \centering
    \includegraphics[width=\linewidth]{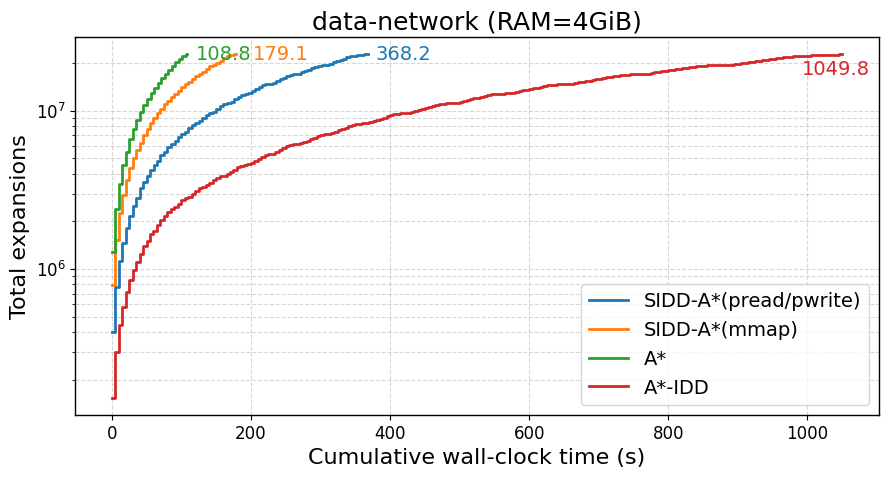}
    \par\vspace{0.4em}
    \includegraphics[width=\linewidth]{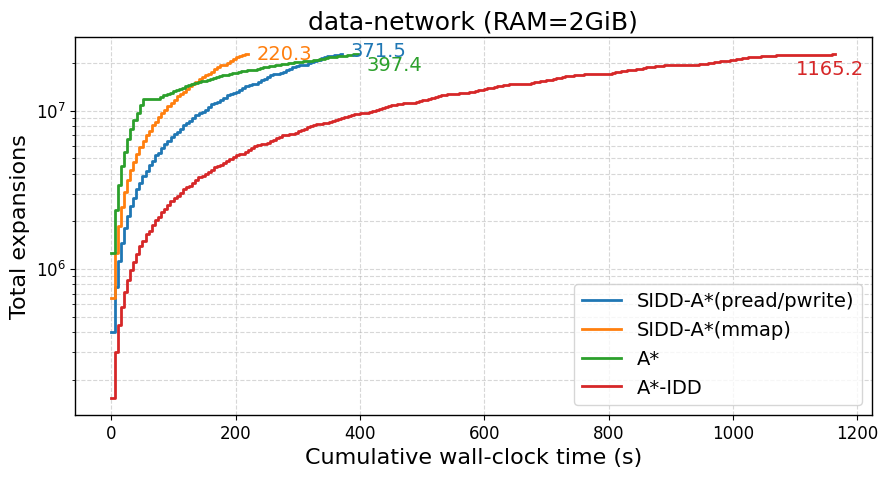}
    \par\vspace{0.4em}
    \includegraphics[width=\linewidth]{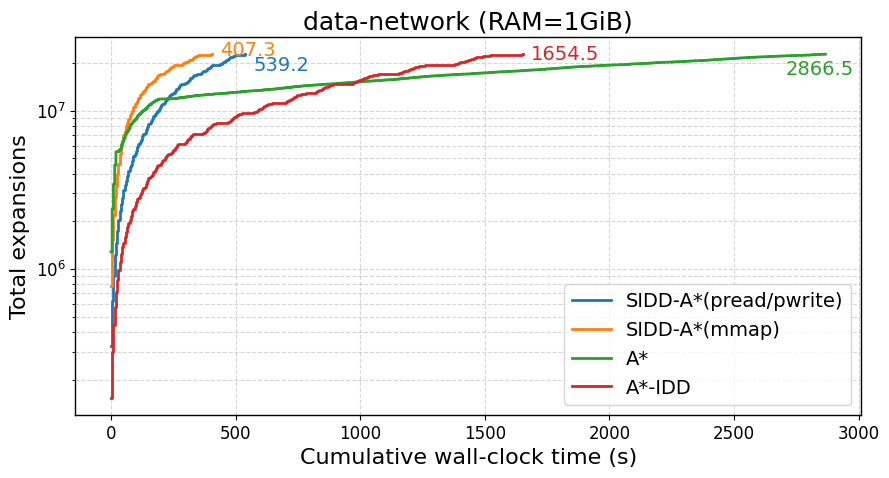}
    \caption{\pddl{data-network-p08}}
    \label{fig:total-data-network-blind-ram-limit}
  \end{subfigure}
  \hfill
  \begin{subfigure}{0.31\textwidth}
    \centering
    \includegraphics[width=\linewidth]{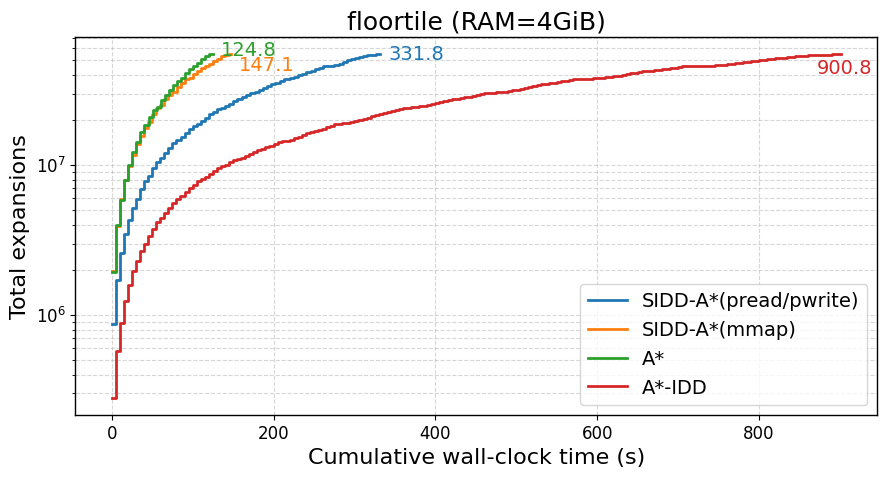}
    \par\vspace{0.4em}
    \includegraphics[width=\linewidth]{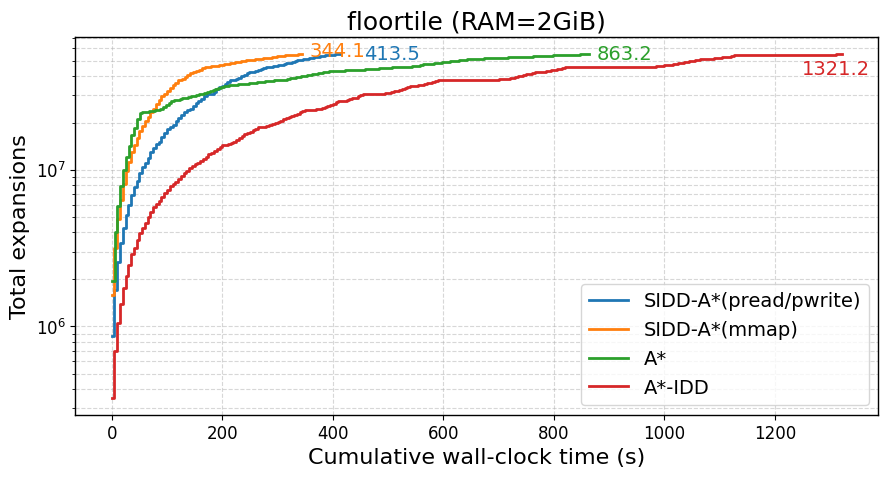}
    \par\vspace{0.4em}
    \includegraphics[width=\linewidth]{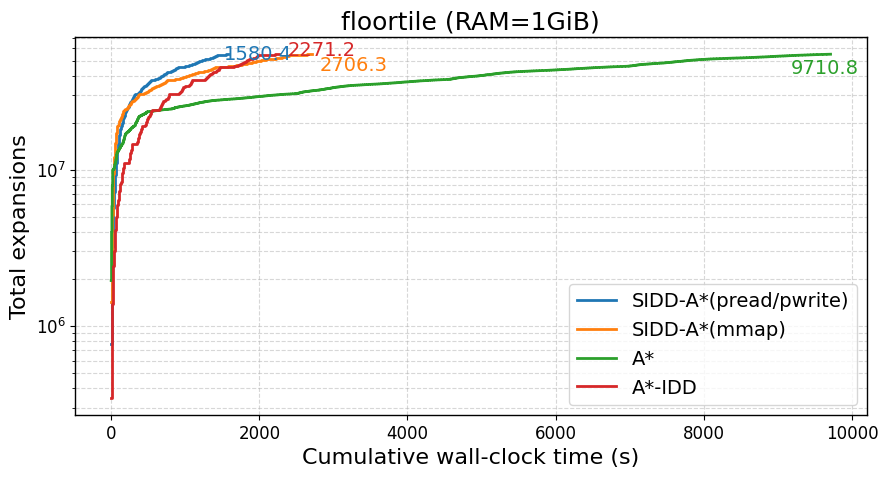}
    \caption{\pddl{floortile-p03}}
    \label{fig:total-floortile-blind-ram-limit}
  \end{subfigure}

  \caption{RAM limitation under the \texttt{blind} heuristic. Each plot shows total expansions as a function of cumulative wall-clock time. Each column shows results for RAM limits of 4\,GiB, 2\,GiB, and 1\,GiB from top to bottom.}
  \label{supp:fig:total-blind-ram-limit}
\end{figure*}

\begin{figure*}[htb]
  \centering

  \begin{subfigure}{0.32\textwidth}
    \centering
    \includegraphics[width=\linewidth]{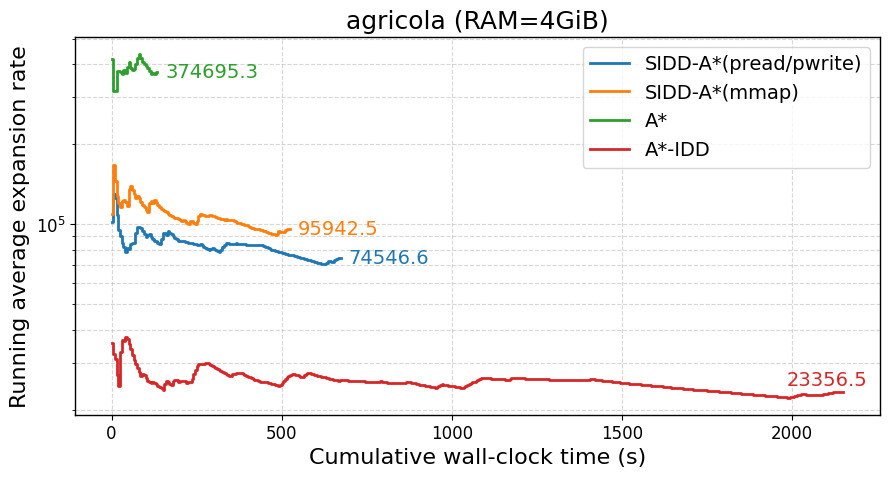}
    \par\vspace{0.4em}
    \includegraphics[width=\linewidth]{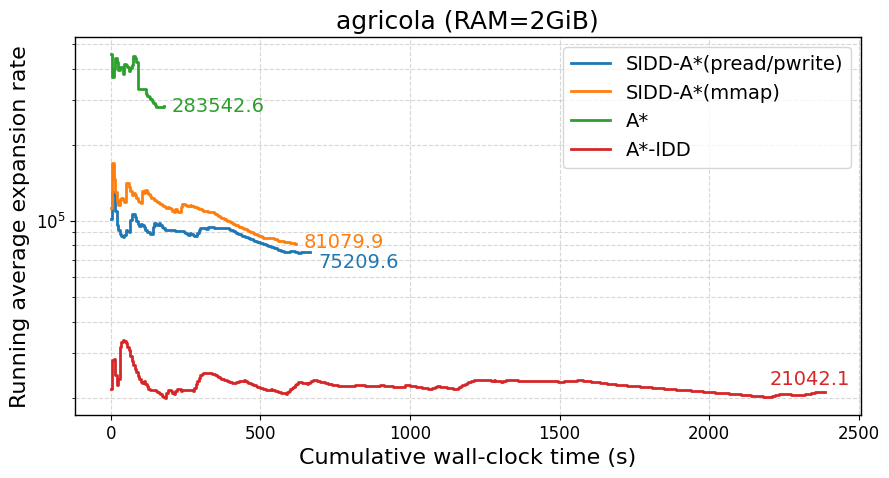}
    \par\vspace{0.4em}
    \includegraphics[width=\linewidth]{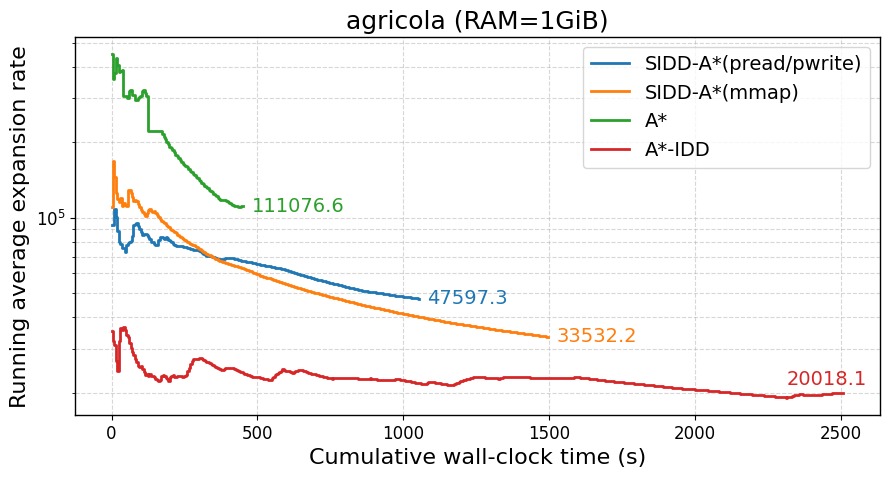}
    \caption{\pddl{agricola-p09}}
    \label{fig:agricola-mands-ram-limit}
  \end{subfigure}
  \hfill
  \begin{subfigure}{0.32\textwidth}
    \centering
    \includegraphics[width=\linewidth]{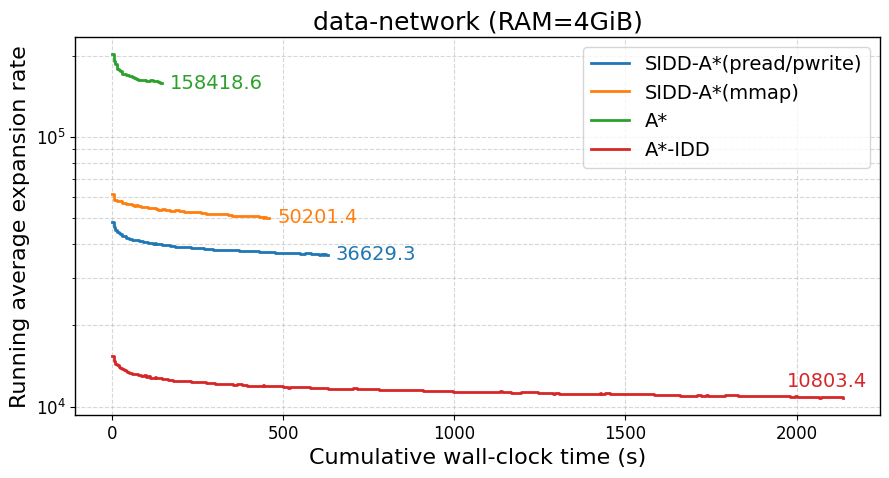}
    \par\vspace{0.4em}
    \includegraphics[width=\linewidth]{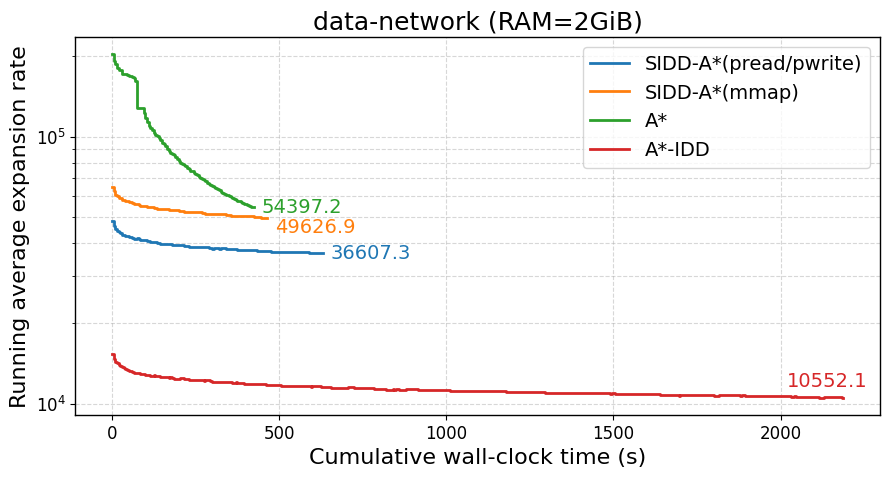}
    \par\vspace{0.4em}
    \includegraphics[width=\linewidth]{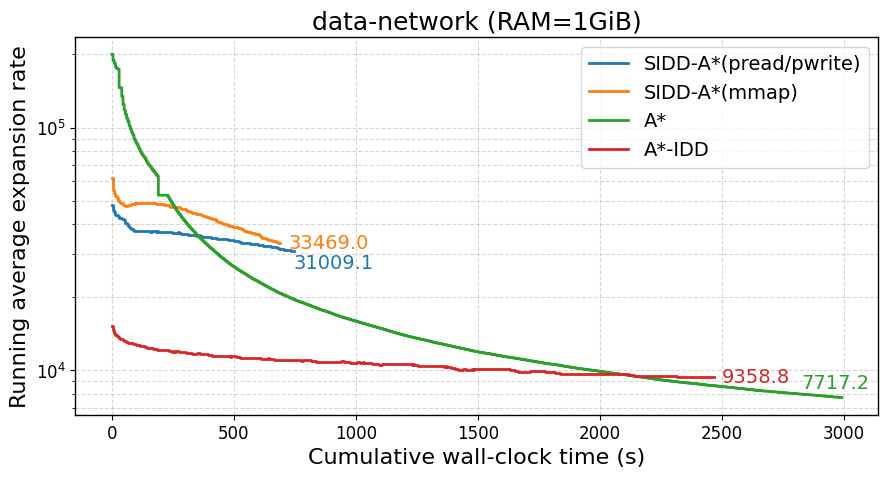}
    \caption{\pddl{data-network-p18}}
    \label{fig:data-network-mands-ram-limit}
  \end{subfigure}
  \hfill
  \begin{subfigure}{0.32\textwidth}
    \centering
    \includegraphics[width=\linewidth]{figures/floortile_4GiB_mands.png}
    \par\vspace{0.4em}
    \includegraphics[width=\linewidth]{figures/floortile_2GiB_mands.png}
    \par\vspace{0.4em}
    \includegraphics[width=\linewidth]{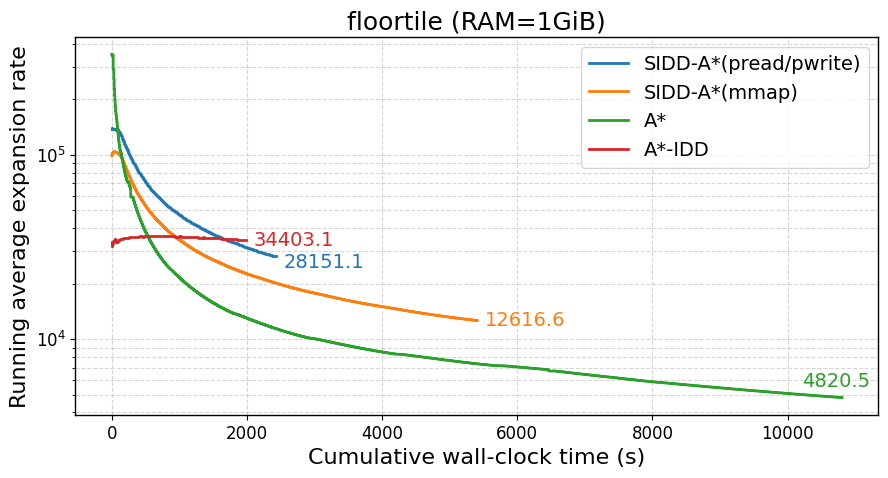}
    \caption{\pddl{floortile-p07}}
    \label{fig:floortile-mands-ram-limit}
  \end{subfigure}

  \caption{RAM limitation under the \texttt{merge-and-shrink} heuristic. Each plot shows running average expansion rate as a function of cumulative wall-clock time. Each column shows results for RAM limits of 4\,GiB, 2\,GiB, and 1\,GiB from top to bottom.}
  \label{supp:fig:merge-and-shrink-ram-limit}
\end{figure*}

\begin{figure*}[htb]
  \centering

  \begin{subfigure}{0.32\textwidth}
    \centering
    \includegraphics[width=\linewidth]{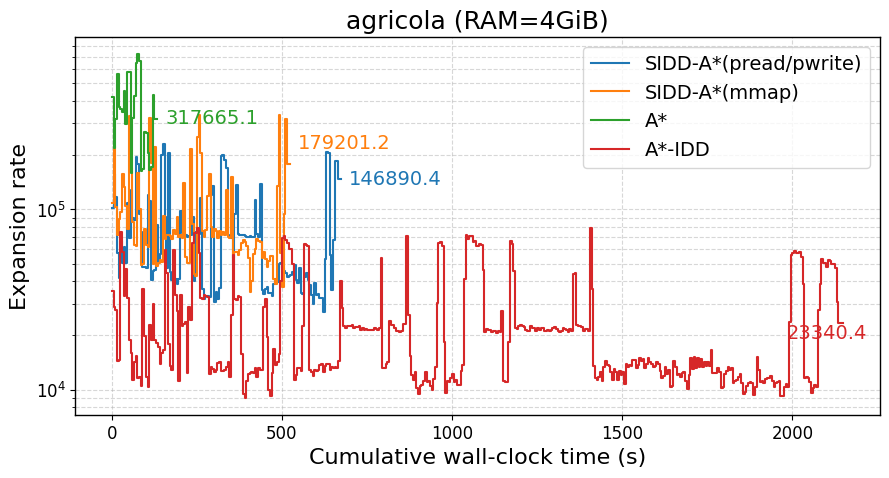}
    \par\vspace{0.4em}
    \includegraphics[width=\linewidth]{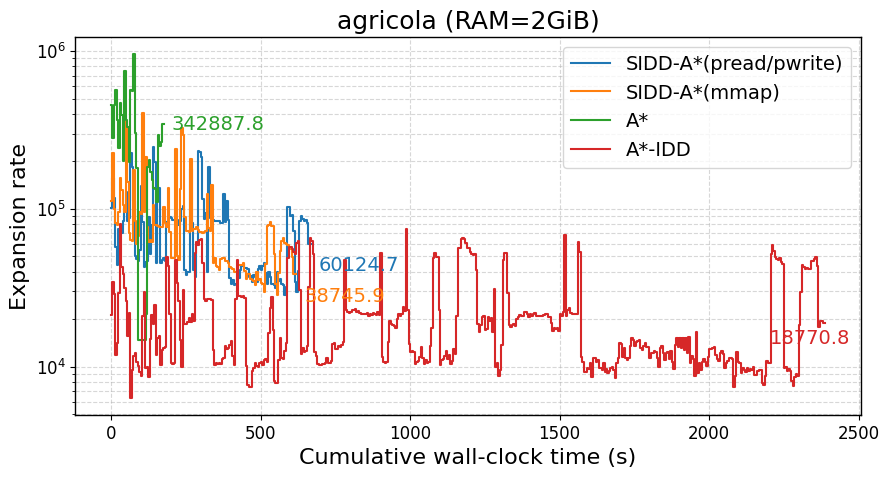}
    \par\vspace{0.4em}
    \includegraphics[width=\linewidth]{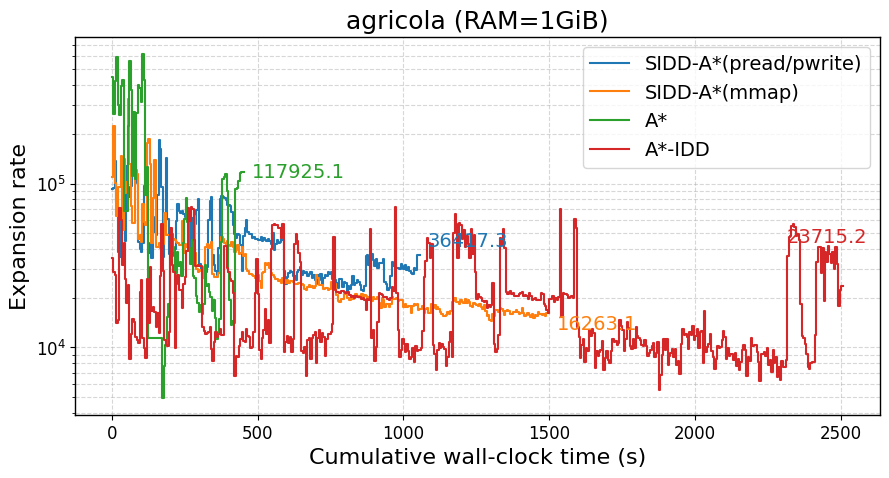}
    \caption{\pddl{agricola-p09}}
    \label{fig:raw-agricola-mands-ram-limit}
  \end{subfigure}
  \hfill
  \begin{subfigure}{0.32\textwidth}
    \centering
    \includegraphics[width=\linewidth]{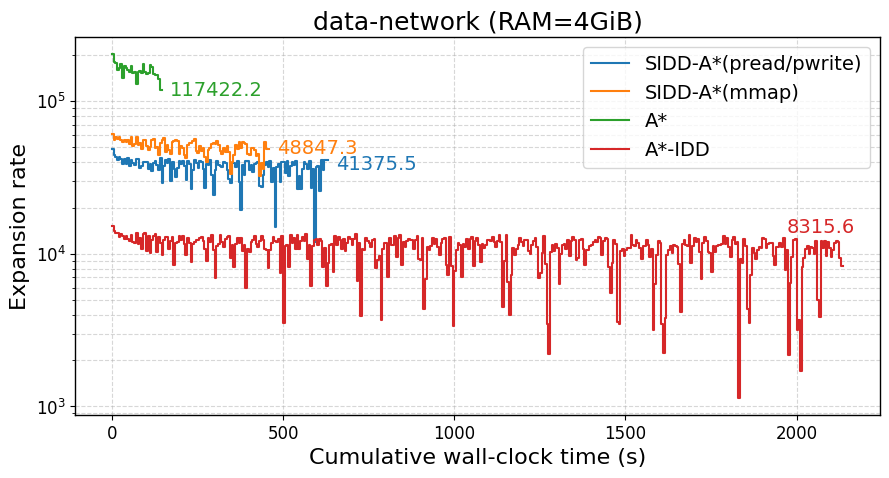}
    \par\vspace{0.4em}
    \includegraphics[width=\linewidth]{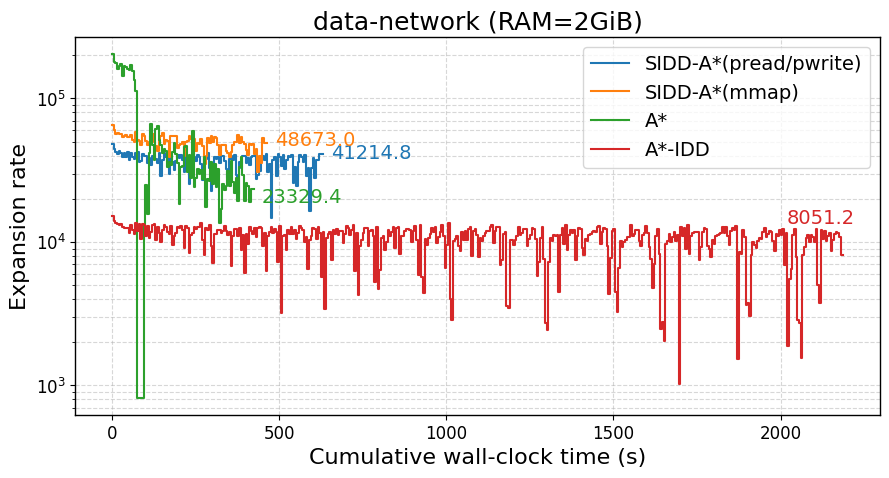}
    \par\vspace{0.4em}
    \includegraphics[width=\linewidth]{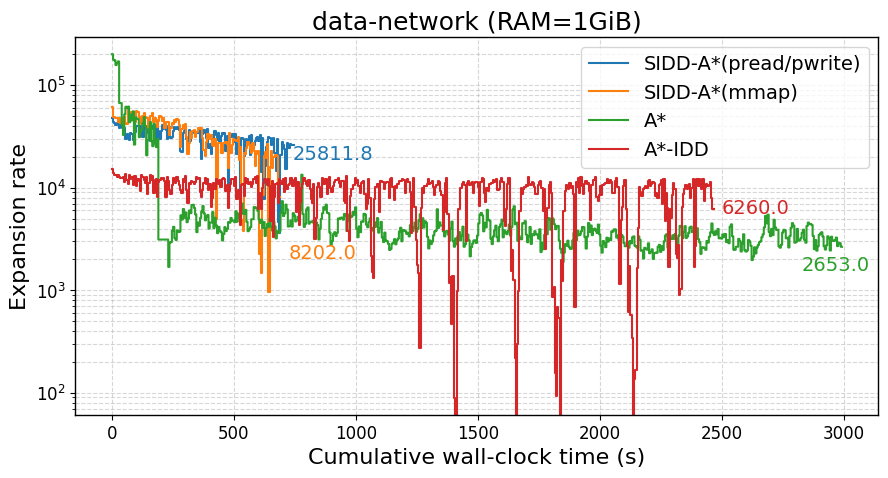}
    \caption{\pddl{data-network-p18}}
    \label{fig:raw-data-network-mands-ram-limit}
  \end{subfigure}
  \hfill
  \begin{subfigure}{0.32\textwidth}
    \centering
    \includegraphics[width=\linewidth]{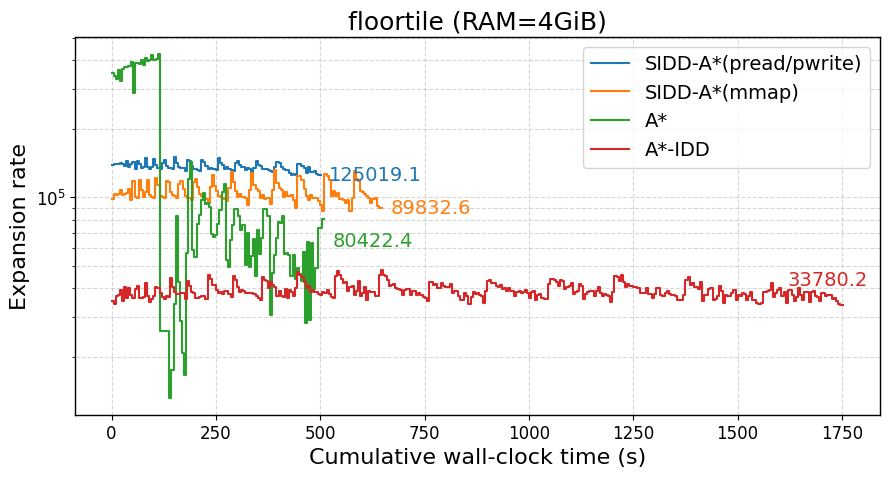}
    \par\vspace{0.4em}
    \includegraphics[width=\linewidth]{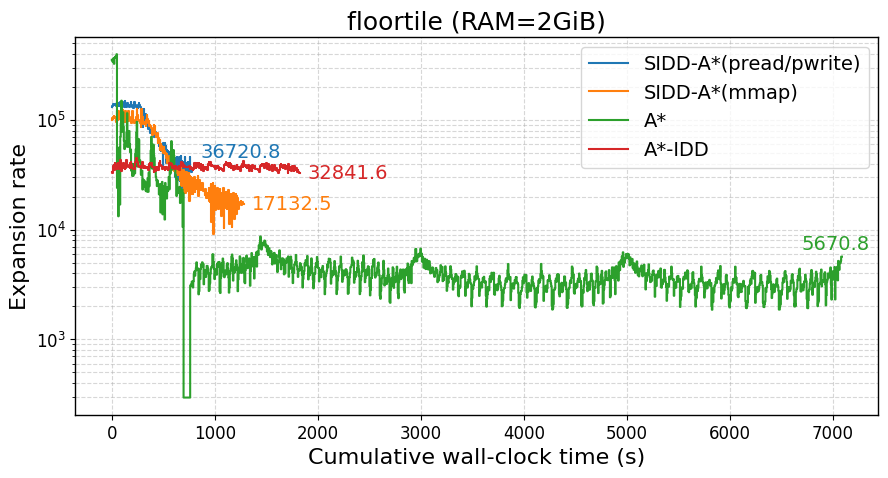}
    \par\vspace{0.4em}
    \includegraphics[width=\linewidth]{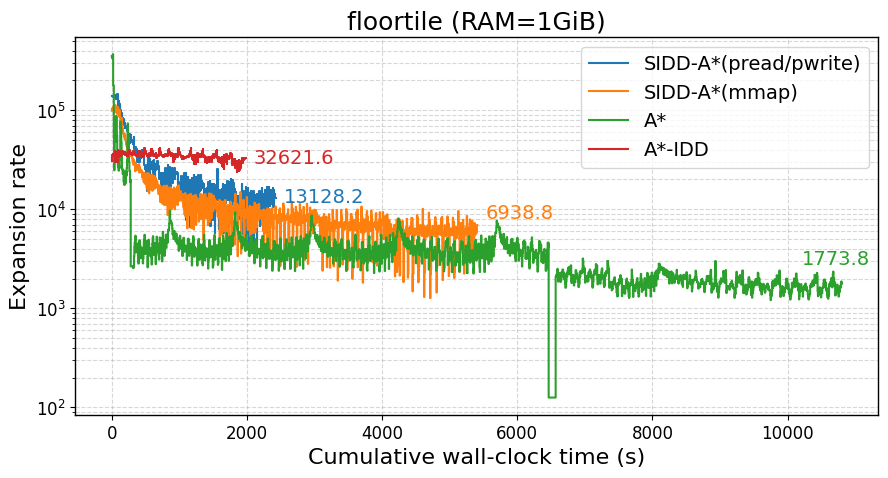}
    \caption{\pddl{floortile-p07}}
    \label{fig:raw-floortile-mands-ram-limit}
  \end{subfigure}

  \caption{RAM limitation under the \texttt{merge-and-shrink} heuristic. Each plot shows expansion rate as a function of cumulative wall-clock time. Each column shows results for RAM limits of 4\,GiB, 2\,GiB, and 1\,GiB from top to bottom.}
  \label{supp:fig:raw-merge-and-shrink-ram-limit}
\end{figure*}

\begin{figure*}[htb]
  \centering

  \begin{subfigure}{0.32\textwidth}
    \centering
    \includegraphics[width=\linewidth]{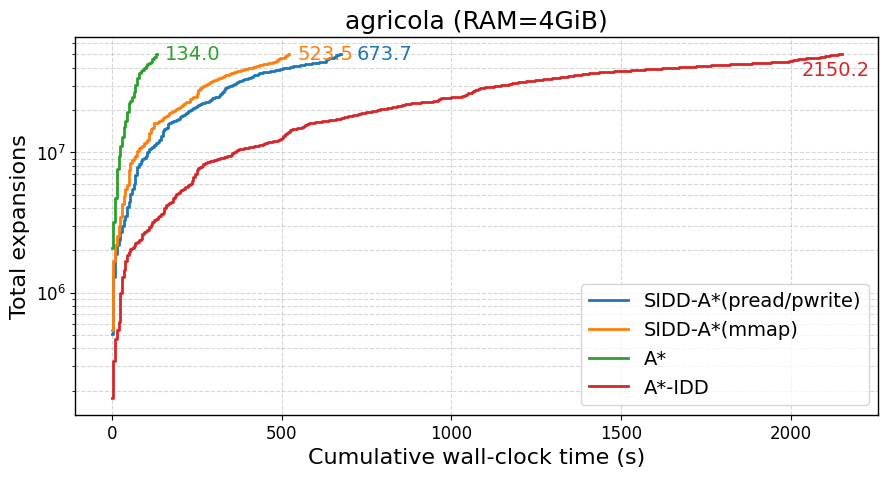}
    \par\vspace{0.4em}
    \includegraphics[width=\linewidth]{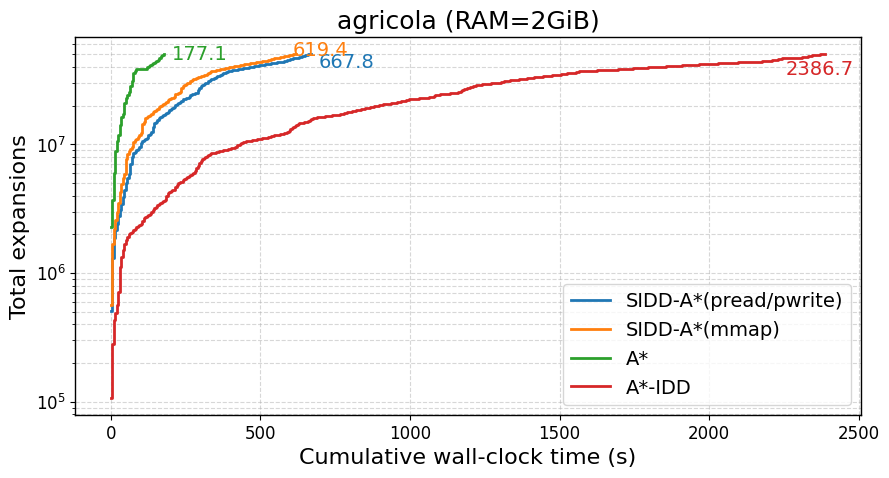}
    \par\vspace{0.4em}
    \includegraphics[width=\linewidth]{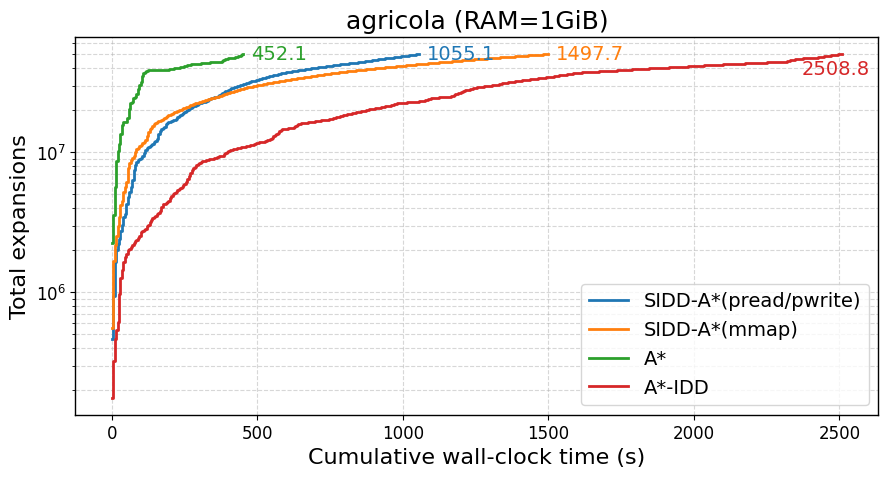}
    \caption{\pddl{agricola-p09}}
    \label{fig:total-agricola-mands-ram-limit}
  \end{subfigure}
  \hfill
  \begin{subfigure}{0.32\textwidth}
    \centering
    \includegraphics[width=\linewidth]{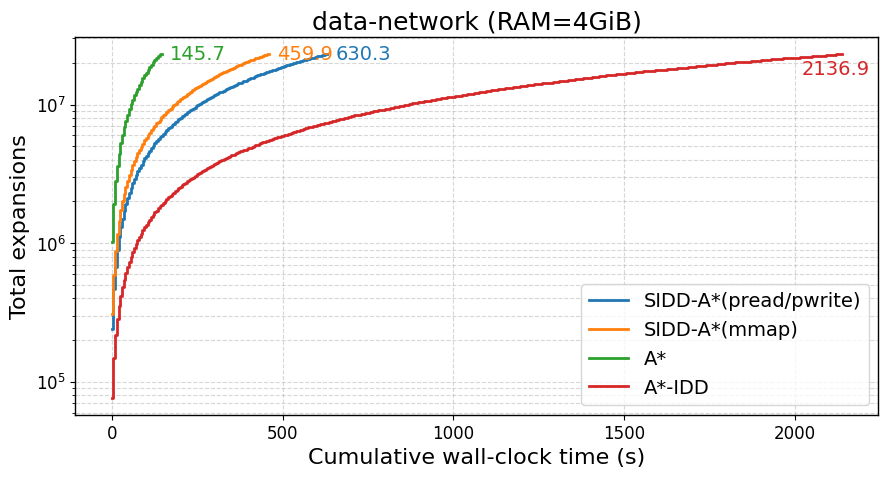}
    \par\vspace{0.4em}
    \includegraphics[width=\linewidth]{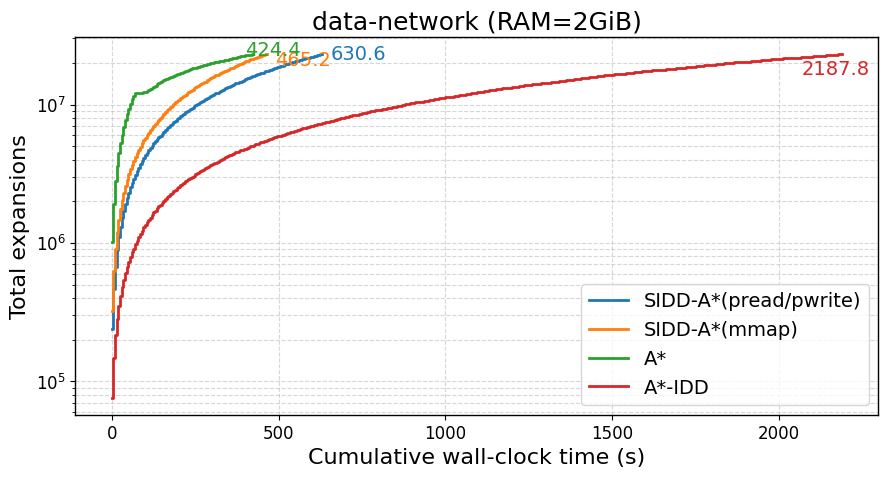}
    \par\vspace{0.4em}
    \includegraphics[width=\linewidth]{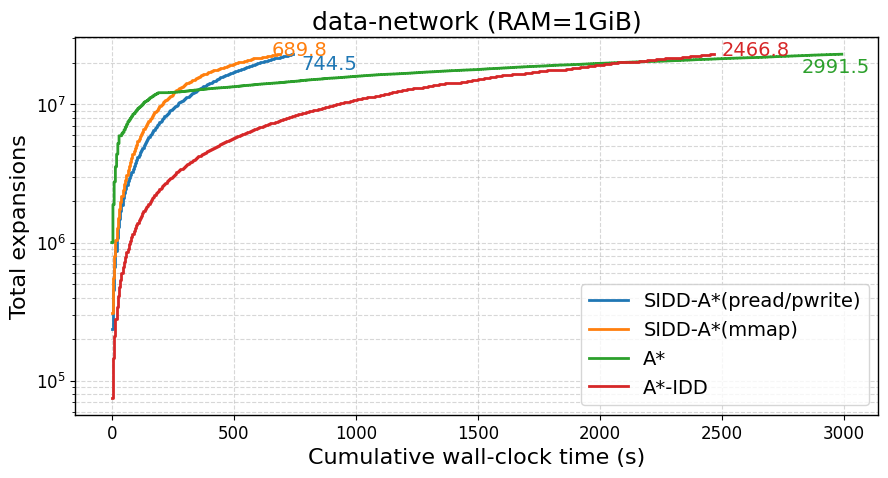}
    \caption{\pddl{data-network-p18}}
    \label{fig:total-data-network-mands-ram-limit}
  \end{subfigure}
  \hfill
  \begin{subfigure}{0.32\textwidth}
    \centering
    \includegraphics[width=\linewidth]{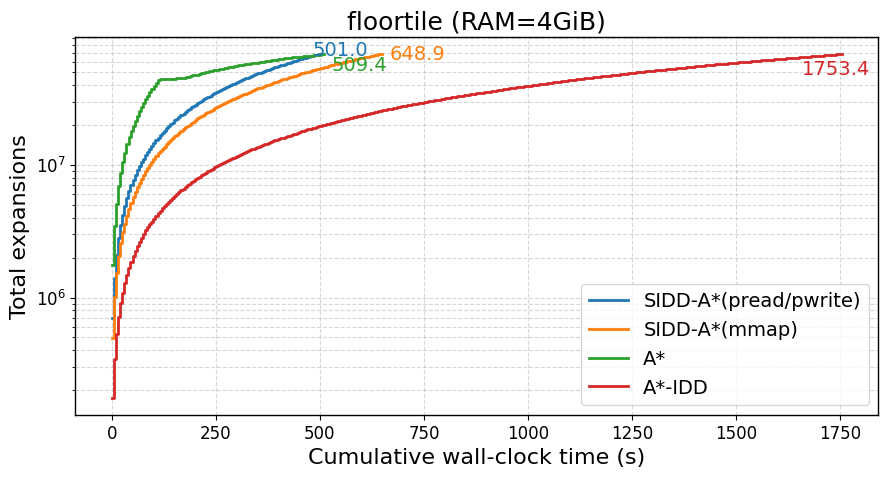}
    \par\vspace{0.4em}
    \includegraphics[width=\linewidth]{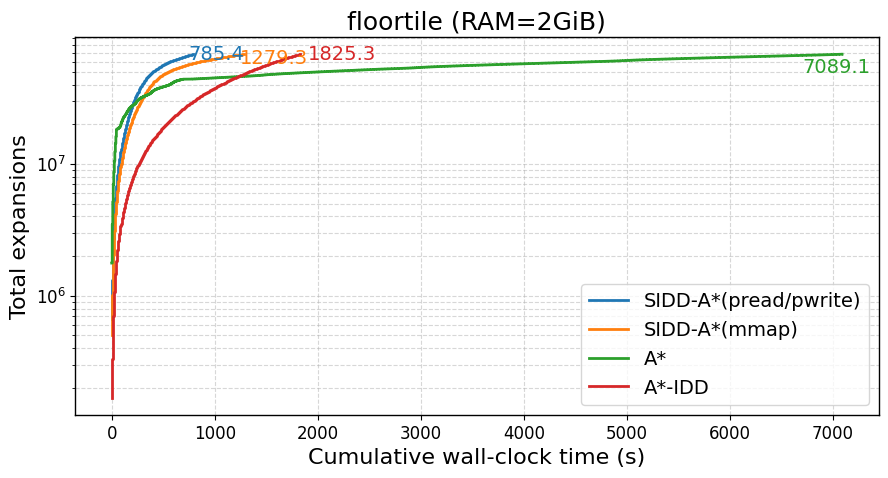}
    \par\vspace{0.4em}
    \includegraphics[width=\linewidth]{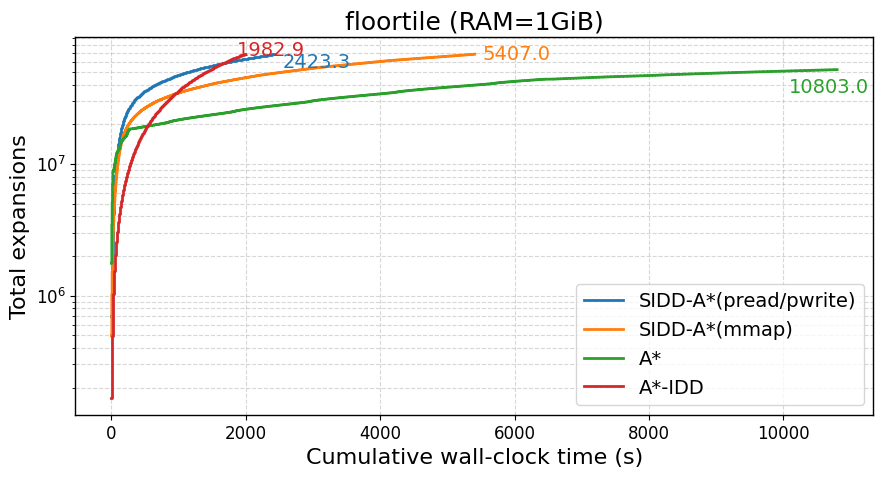}
    \caption{\pddl{floortile-p07}}
    \label{fig:total-floortile-mands-ram-limit}
  \end{subfigure}

  \caption{RAM limitation under the \texttt{merge-and-shrink} heuristic. Each plot shows total expansions as a function of cumulative wall-clock time. Each column shows results for RAM limits of 4\,GiB, 2\,GiB, and 1\,GiB from top to bottom.}
  \label{supp:fig:total-merge-and-shrink-ram-limit}
\end{figure*}

\fi

\end{document}